\documentclass{article} %
\usepackage{iclr2027_conference,times}

\usepackage{amsmath,amsfonts,bm}

\def\eqref#1{equation~\ref{#1}}

\def\1{\bm{1}}

\DeclareMathAlphabet{\mathsfit}{\encodingdefault}{\sfdefault}{m}{sl}
\SetMathAlphabet{\mathsfit}{bold}{\encodingdefault}{\sfdefault}{bx}{n}

\usepackage{algorithm}
\usepackage{algpseudocode}

\usepackage{hyperref}
\usepackage{url}
\usepackage{wrapfig}
\usepackage{enumitem}
\usepackage[most]{tcolorbox}
\usepackage{xcolor}
\usepackage{amsmath}
\usepackage{minitoc}

\definecolor{boxblue}{HTML}{2449A4}
\definecolor{boxfill}{HTML}{EEF4FF}

\usepackage{booktabs}
\usepackage{multirow}
\usepackage{tabularx}
\usepackage{array}
\usepackage{adjustbox}
\usepackage[table]{xcolor}

\definecolor{bestblue}{RGB}{0,128,128}
\definecolor{secondorange}{RGB}{190,120,45}
\definecolor{sectiongray}{RGB}{235,235,235}

\definecolor{toolblue}{HTML}{3B6085}
\definecolor{toolgreen}{HTML}{437C38}
\definecolor{toolpurple}{HTML}{73568C}

\newtcolorbox{highlightbox}{
  colback=boxfill,
  colframe=boxblue,
  boxrule=0.8pt,
  arc=4pt,
  left=12pt,
  right=12pt,
  top=9pt,
  bottom=9pt,
  before skip=10pt,
  after skip=10pt
}

\definecolor{boxgray}{RGB}{130,130,130}
\definecolor{boxbg}{RGB}{248,248,248}

\usepackage[most]{tcolorbox}

\newtcolorbox{instructionbox}[1]{
    enhanced,
    breakable,
    title={#1},
    colback=gray!5,
    colframe=gray,
    colbacktitle=gray,
    coltitle=white,
    fonttitle=\normalsize\bfseries,
    boxrule=0.8pt,
    rounded corners=all,
    arc=3mm,
    outer arc=3mm,
    left=4mm,
    right=4mm,
    top=3mm,
    bottom=3mm,
    toptitle=1mm,
    bottomtitle=1mm,
    titlerule=0pt
}

\title{How Harnesses Reshape Agentic Search: Training Stateless Large Reasoning Models for Stateful Search }

\author{\normalfont\normalsize\bfseries Jiayuan Wang$^{1,*}$, Kaikai Wang$^{1,*}$, Weijun Huang$^{4}$, Jingkai Shao$^{1}$, Chen Shu$^{1}$ \\
\normalfont\normalsize\bfseries Xiaowen Zhang$^{1}$, Sutong Wang$^{1,2}$, Qingshi Zhou$^{3}$, Chen Yao$^{1,\dagger}$, Shiping Wang$^{4,\dagger}$ \\[3pt]
\normalfont\normalsize $^{1}$Ant Group \quad $^{2}$Northwest Polytechnical University \\
\normalfont\normalsize $^{3}$Shanghai Jiao Tong University \quad $^{4}$Fuzhou University \\[2pt]
\normalfont\small $^{*}$Equal contribution. \quad $^{\dagger}$Co-corresponding authors.}

\iclrfinalcopy %
\begin{document}

\maketitle
\pagestyle{fancy}
\fancyhead{}
\fancyhead[L]{Preprint}
\renewcommand{\headrulewidth}{0.4pt}
\thispagestyle{fancy}

\begin{abstract}
Agentic search requires reasoning over evidence accumulated across interactions, yet the underlying language model is stateless across invocations. Existing approaches typically bridge this gap by carrying interaction histories across turns and applying post-training techniques. However, growing contexts can obscure earlier evidence and force the model to reconstruct search progress at each decision. Moreover, when supervision relies solely on final outcomes, training provides limited corrective feedback, leaving intermediate errors unaddressed and potentially reinforcing flawed reasoning or redundant actions.
We introduce H-Search, a framework that addresses these challenges through the joint design of a stateful search harness and decision-level policy training. H-Search organizes multi-turn search into self-contained decisions: each model invocation conditions on the current structured state, with cross-turn dependencies preserved through explicit state transitions. This separation makes the same state-conditioned decision the basic unit of both search execution and policy supervision. To train within this structure, we develop harness-grounded on-policy self-distillation, where a frozen copy of the initial model uses privileged information about evidence coverage and termination readiness to provide dense token-level guidance along student-generated trajectories. The resulting policy operates within the same harness at inference without privileged information or teacher access. Across five open-domain benchmarks over Wikipedia, H-Search with a 27B backbone achieves 79.8\% average recall, outperforming the evaluated same-backbone baselines by at least 11.2 percentage points. These results demonstrate the potential of coupling explicit search-state management with supervision at the level of individual decisions.

\end{abstract}

\begin{figure}[h]
\centering
\includegraphics[width=\linewidth]{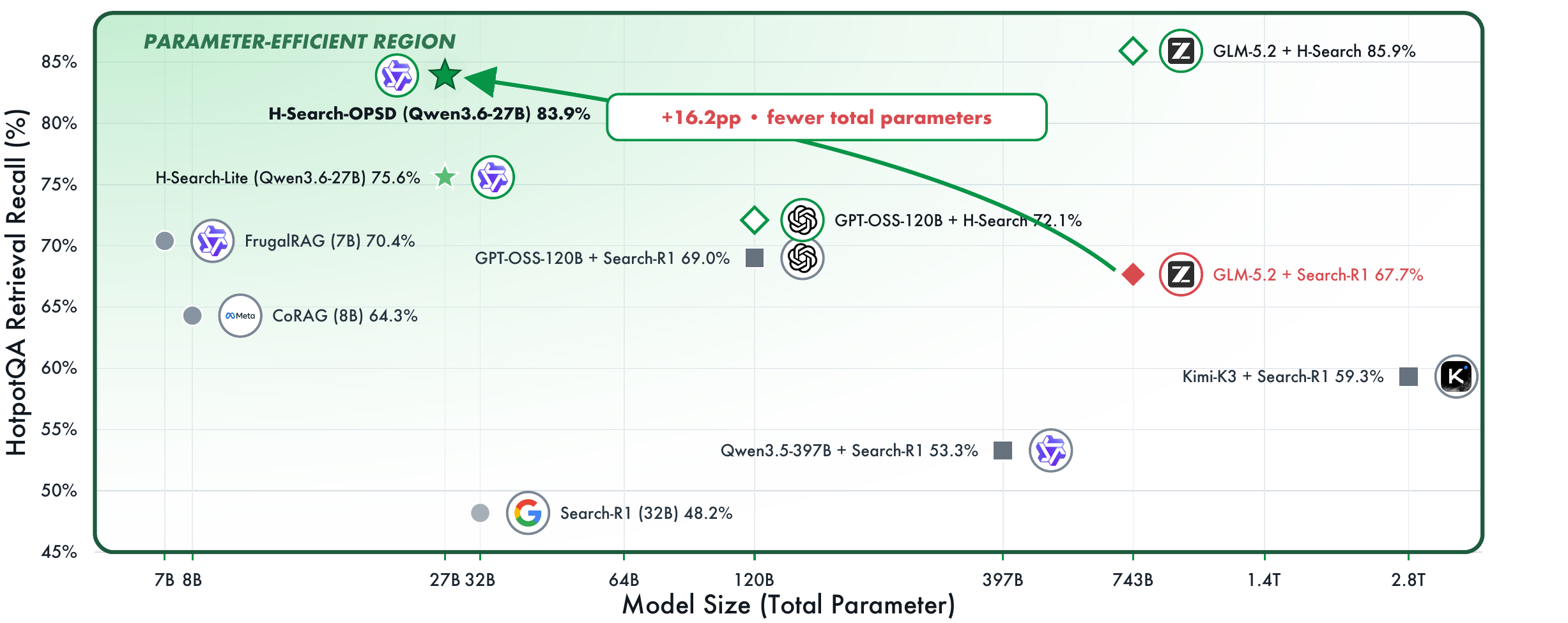}
\caption{\textbf{Parameter efficiency on HotpotQA.} H-Search with Qwen3.6-27B achieves 83.9\% performance, outperforming GLM-5.2 with Search-R1 by 16.2 percentage points while using 27.5× fewer total model parameters. The results demonstrate that H-Search enables a relatively small model to reach frontier-level performance, substantially reducing its dependence on model scale.}
\label{fig:intro_acc}
\end{figure}

\section{Introduction}
In recent years, Large Reasoning Models~(LRMs) have made substantial progress in complex reasoning and instruction-following tasks~\citep{li2025search,wang2026multi,guo2025deepseek}. By integrating search engines, knowledge bases, and document-reading tools, these models have evolved beyond static question answering systems that rely primarily on parametric knowledge into search agents capable of proactively acquiring external information~\citep{li2026webthinker}. Unlike conventional retrieval-augmented approaches, which retrieve information once and then generate an answer, agentic search enables models to decompose user requests, formulate queries, and continually adapt subsequent actions based on newly acquired evidence over multiple rounds of interaction~\citep{wu2026webdancer,press2023measuring}. This interleaving of reasoning and search has become an important foundation for knowledge-intensive tasks, including open-domain question answering, complex information retrieval, and deep research. It also enables models to produce more accurate and traceable answers when their internal knowledge is incomplete or when a question requires evidence from multiple reasoning steps~\citep{nakano2021webgpt}.

Simply providing an LRM with access to search tools does not, in itself, yield a reliable search agent~\citep{song2025r1}. Common failures in multi-turn search—such as losing previously gathered evidence, making ineffective tool calls, and terminating either prematurely or too late—are often attributed to insufficient reasoning capabilities~\citep{liu2024lost}. Accordingly, existing studies have primarily sought to improve performance through better retrievers, reinforcement learning, or other post-training strategies~\citep{jin2025empirical}. Although recent work has begun to incorporate shared memory, modular control, and training objectives designed for long-horizon interaction~\citep{li2026se}, these execution mechanisms are typically treated as fixed system implementations rather than as core variables that shape the learning process. In practice, large language models do not autonomously maintain persistent state during interaction. Instead, the external harness determines how the search history, candidate documents, tool feedback, and remaining budget are selected, organized, and presented to the model at each decision step~\citep{packer2023memgpt}.
\begin{highlightbox}
\textbf{Harness Matters.} {The harness constrains the available actions, state transitions, and termination conditions, thereby directly shaping both the distribution of states encountered by the model during training and the supervision signals it receives. Without explicitly accounting for this layer, it is difficult to determine whether performance improvements arise from the model itself or from the design of the search agent in which it operates.}
\end{highlightbox}

\begin{figure}[t]
\centering
\includegraphics[width=\linewidth]{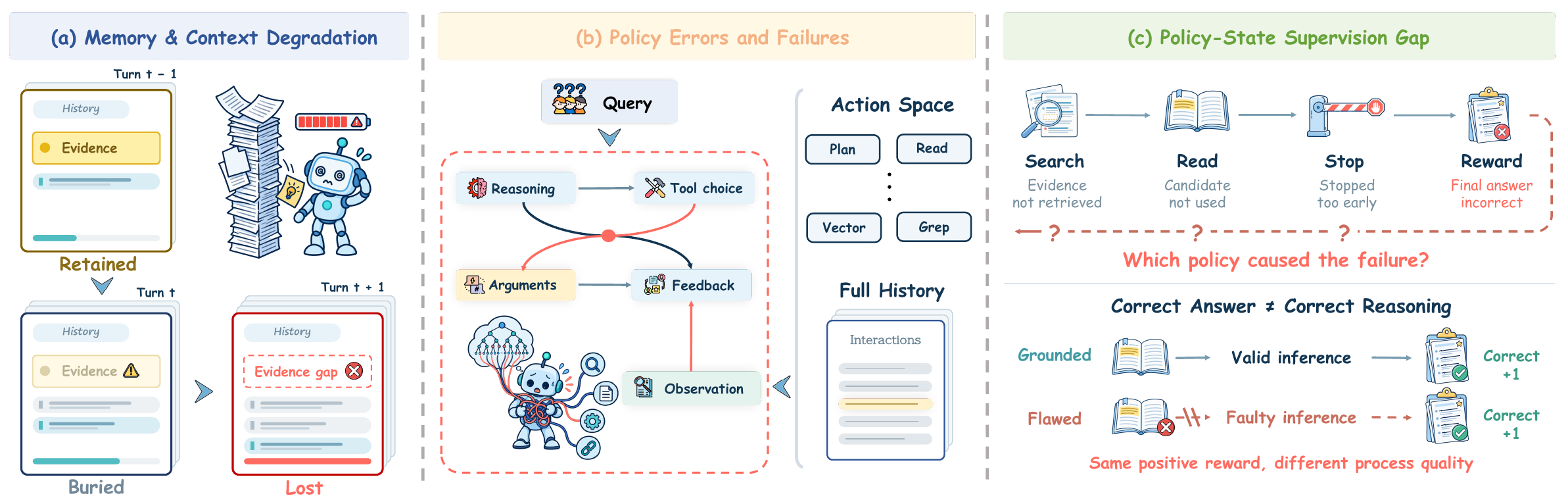}
\caption{\textbf{Three challenges in agentic search.} (a) Previously acquired evidence can become buried or lost as the interaction history grows. (b) Selecting tools and arguments requires reconstructing search progress from the available context, increasing the risk of ineffective actions. (c) Final-answer rewards provide insufficient guidance for intermediate decisions: an incorrect answer may arise from different search failures, while a correct answer may conceal flawed reasoning.}
\label{fig:frame}
\end{figure}

From this perspective, reliable multi-turn search faces three interrelated challenges (as illustrated in Figure \ref{fig:frame}): 
\textbf{\romannumeral 1) Memory and Context Degradation:} As queries, candidate documents, and tool feedback accumulate, evidence collected earlier may become buried in redundant information or lost during context truncation and compression, disrupting the evidence chain needed for subsequent decisions. \textbf{\romannumeral 2) Policy Errors and Failures:} The model must not only select tools and their arguments but also reconstruct search progress from a mixed interaction history, distinguishing evidence already acquired from candidates awaiting examination and information gaps that remain unresolved. 
This increases the risk of redundant retrieval, ineffective tool calls, and premature termination decisions.
\textbf{\romannumeral 3) Policy-State Supervision Gap:} The same incorrect answer may result from missing evidence, unused candidates, or premature termination, while a correct answer may still involve faulty inferences, redundant searches, or unsupported conclusions. Assigning rewards solely based on the final answer therefore makes it difficult to identify which decisions require correction and may reinforce unreliable behavior that succeeds by chance. Together, these challenges raise a central question: \textit{how can the information available to the policy, its permitted actions, and its training feedback remain aligned with the evolving evidence state and the decisions that state requires?}

To address these challenges, we propose H-Search, a framework for training stateless large reasoning models to perform stateful search. H-Search separates persistent search state from model inference, organizing multi-turn interaction as a sequence of self-contained decisions conditioned on the current state. The harness preserves the consequences of earlier actions and constrains the available choices, allowing the model to focus on what to do next. This decomposition makes it possible to supervise a search trajectory at the granularity of individual decisions. 
We jointly design the harness and its training interface around these state-conditioned decisions. The harness explicitly maintains search progress and defines the context and admissible actions for each policy invocation, making the same decision the basic unit of both execution and supervision. Building on this interface, we adapt on-policy self-distillation to provide dense token-level guidance along student-generated trajectories. At each visited state, a frozen teacher receives the student’s decision context augmented with privileged feedback about evidence coverage, candidate relevance, and termination readiness. This couples policy supervision to the evidence state maintained by the harness and the decision the student must make within it.
Our main contributions are threefold:
\begin{itemize}[leftmargin=*]
    \item \textbf{Joint Harness–Training Design:} We introduce H-Search, a framework that jointly designs a stateful search harness and decision-level policy training, using independent, state-conditioned decisions as the shared unit of execution and supervision.
    \item \textbf{Harness-Grounded Policy Supervision:} We couple explicit evidence-state management and state-dependent action constraints with on-policy self-distillation, where privileged evidence-state feedback guides the student’s decisions along its own search trajectories.
    \item \textbf{Empirical Evaluation: } Experiments on five open-domain benchmarks show that H-Search enables a 27B model to approach the retrieval performance of substantially larger models.
\end{itemize}

\section{Related Work}
\label{Related Work}
\paragraph{Agentic Search.} Early work on knowledge-intensive question answering primarily follows a retrieve-then-generate paradigm. Dense retrievers such as DPR~\citep{karpukhin-etal-2020-dense} learn to retrieve relevant passages from large corpora, while RAG~\citep{lewis2020retrieval} and Fusion-in-Decoder~\citep{izacard-grave-2021-leveraging} condition language generation on retrieved documents. 
These methods typically perform retrieval using the original input and therefore provide limited support for tasks in which later information needs depend on earlier findings. Subsequent work introduces more adaptive retrieval strategies. IRCoT~\citep{trivedi2023interleaving} interleaves retrieval with chain-of-thought reasoning, FLARE~\citep{jiang-etal-2023-active} triggers retrieval according to uncertainty during generation, and Self-RAG~\citep{asai2024self} learns to retrieve and critique retrieved evidence through reflection tokens. ReAct~\citep{yao2022react} establishes the reasoning–action–observation interaction pattern; Search-o1~\citep{li2025search} integrates on-demand search into long-form reasoning; and DeepResearcher~\citep{zheng-etal-2025-deepresearcher} extends this paradigm to live web search and browsing. 
These studies mainly improve the model or workflow under a particular agent architecture. In contrast, we treat the Harness itself as part of the effective decision process and study how its state representation, action constraints, and control flow reshape the behavior learned by a stateless LRM.
\paragraph{Post-training for Search.} Post-training a search agent differs from standard language-model fine-tuning because each training example contains multiple model decisions interleaved with non-differentiable tool responses, while task success is often observed only after the complete trajectory.
Existing approaches can therefore be distinguished by how supervision is obtained and assigned to intermediate search decisions. Outcome-based RL instead allows the policy to explore within the search environment: Search-R1~\citep{jin2025search} and ReSearch~\citep{chen2025research} optimize interleaved reasoning and retrieval using final-answer rewards, while DeepResearcher~\citep{zheng-etal-2025-deepresearcher} and ASearcher~\citep{gao2025beyond} extend end-to-end RL to real-web and long-horizon settings. However, assigning a single terminal reward to an entire trajectory provides limited guidance about which intermediate search actions are useful. Recent work addresses this issue through finer-grained credit assignment: AgentFlow optimizes actions collected from in-system rollouts~\citep{li2026flow}, MemSearcher~\citep{yuan-etal-2026-memsearcher} propagates trajectory-level advantages across turns with distinct contexts. 
These methods have substantially advanced search. However, trajectory-level rewards remain inherently sparse, while training with a Harness requires fine-grained adjustment of intermediate decisions throughout the trajectory, making it difficult to assign credit to individual actions.

\section{Methodology}
\label{Methodology}

\begin{figure}[t]
\centering
\includegraphics[width=\linewidth]{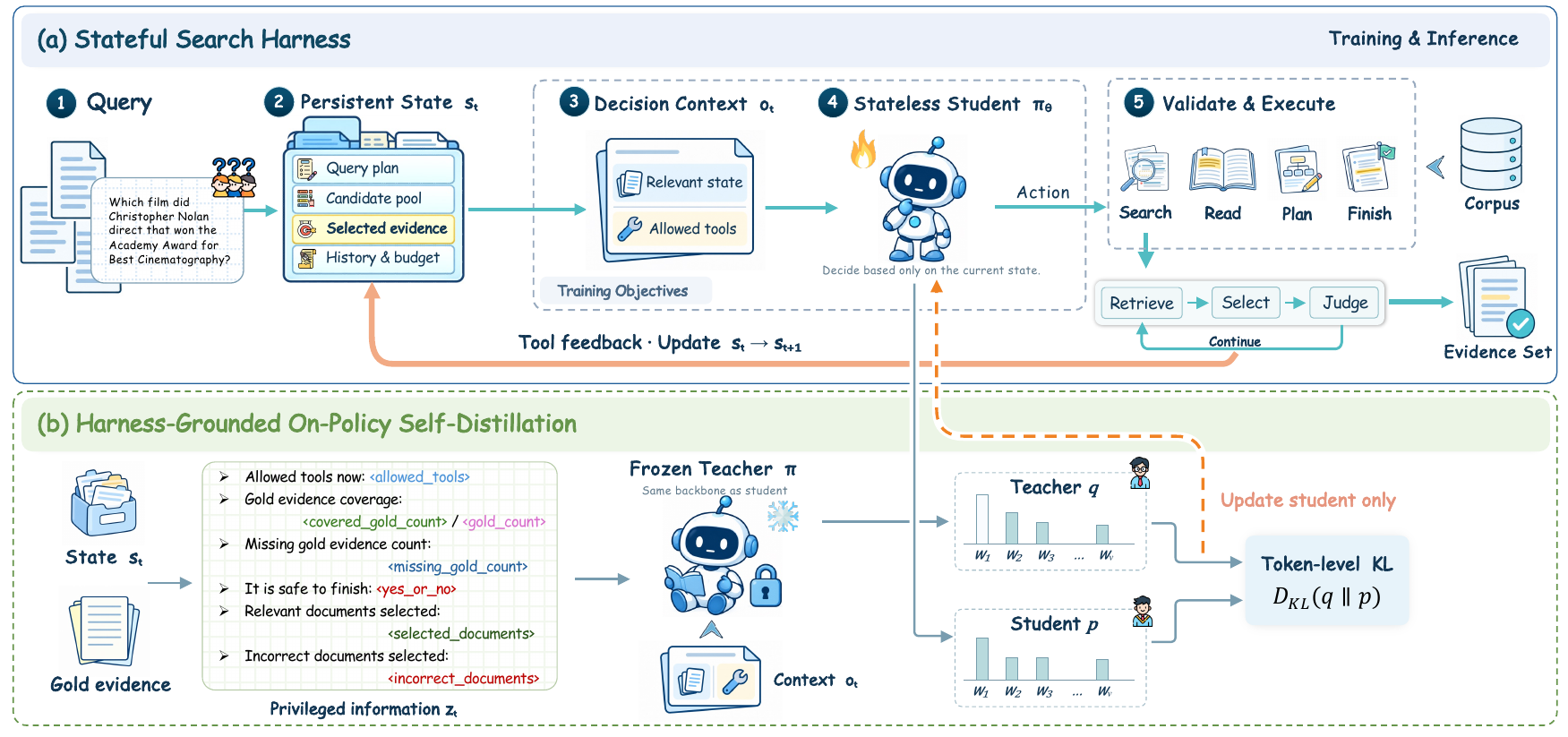}
\caption{\textbf{Overview of H-Search.} (a) A stateful harness manages persistent search state and action execution, while a stateless policy acts on the current context. (b) The same state, combined with privileged teacher feedback for OPSD, jointly connecting harness design and policy training.}
\label{fig:framework}
\end{figure}

We present H-Search, as illustrated in figure~\ref{fig:framework}, which couples a stateful search harness with fine-grained on-policy trajectory distillation. We first formulate the search problem, then describe the harness that structures state-conditioned decisions and the methods based on this decision structure.
\subsection{Problem Formulation}
Given a query $q$ and an external corpus $\mathcal{D}$, the goal of the agent is to identify, through multiple rounds of tool interaction, an evidence set $E$ that supports answering the query. Following prior work~\citep{he-etal-2025-pasa,shi2026iterative}, we formulate
search as a finite-horizon sequential decision process, in which the model alternates among retrieval, reading, and evidence selection until it actively terminates the search or exhausts the search budget.

We model search as a finite-horizon interactive process orchestrated by the harness framework. At round $t$, the harness maintains a search state $s_t$ that contains the query plan, candidate documents, search history, and remaining budget. A state-rendering function $R$ transforms $s_t$ into the model-visible context $o_t$, after which a stateless LRM policy $\pi_\theta$ generates a raw action $a_t$:
\begin{equation}
    o_t = R(s_t),
    \qquad
    a_t \sim \pi_\theta(\cdot \mid o_t),
\end{equation}
where $\theta$ denotes the trainable model parameters. All information required for cross-round decision making is reintroduced by the Harness through $o_t$.

Let $\mathcal{Q}$ denote the query distribution, $G(q)\subseteq\mathcal{D}$ the reference evidence set for query $q$, and $U(E,G(q))$ a utility function that measures the quality of the final evidence set. The objective is to learn a policy $\pi_\theta$ that, subject to a given search budget, selects evidence that covers the information required by the query while minimizing the inclusion of irrelevant documents:
\begin{equation}
    \theta^\star
    =
    \arg\max_\theta
    \mathbb{E}_{\substack{
        q\sim\mathcal{Q},
        \tau\sim\pi_\theta
    }}
    \left[
        U\bigl(E,G(q)\bigr)
    \right],
\end{equation}
here, $\tau\sim\pi_\theta$ denotes an interaction trajectory jointly induced by
the model policy and the search environment.

\subsection{Stateful Search Harness}

\label{Stateful Search Harness}
Although search tasks inherently involve dependencies across interaction turns, the language model responsible for decision-making does not itself maintain persistent state. At each invocation, the model selects an action based solely on the context $o_t$ reconstructed and supplied by the harness. Consequently, whether the search history is preserved, how candidate documents are organized, and which information is exposed to the model in the next decision step are not intrinsic properties of the model policy. Instead, they are determined by the harness’s state representation and context-rendering mechanisms. Based on this observation, H-Search is designed as an explicit state-management interface between the model and the search environment, thereby eliminating the need for the model to reconstruct cross-turn information from unstructured interaction histories.

\paragraph{Policy–State Decoupling.} At search turn \(t\), H-Search maintains the complete state \(s_t\) and transforms it into the model-visible context \(o_t\) through a state-rendering function \(R\):
\begin{equation}
o_t = R(s_t) = R\bigl({q}, \mathcal{M}_t, \mathcal{H}_t,\mathcal{A}_t\bigr),
\end{equation}
here, $\mathcal{M}_t$, $\mathcal{H}_t$, and $\mathcal{A}_t$ denote the current evidence state, the search and interaction history, and the set of actions available under the current state, respectively. While the complete state is persistently maintained by the harness, $R$ exposes only the information relevant to the current decision by selectively compressing and organizing the state. The model therefore need not reproduce tool outputs during generation or maintain long-term memory on its own; it only needs to select the next action based on $o_t$. This decoupling allows policy learning to focus on conditional decision-making, without requiring the policy to learn how to compress interaction histories, retain retrieved documents, or recover task progress. Moreover, the consistent state interface across turns provides unified state semantics for subsequent policy optimization over interaction trajectories.
\paragraph{Harness-Constrained Agentic Search.}
Building on the structured and explicit search state, H-Search provides the policy with a unified interface for interacting with the environment, enabling it to acquire external information, update the evidence set, and determine whether to continue searching. The model can thus adaptively select the search granularity and direction of exploration based on the currently available evidence, without following a predefined retrieval trajectory. The Harness constrains this autonomous search process through a state-dependent action space. Let $\mathcal{A}$ denote the complete action space. The set of executable actions at step $t$ is given by:
\begin{equation}
    \mathcal{A}_t = \Gamma(s_t) \subseteq \mathcal{A},
\end{equation}
where $\Gamma$ determines the valid actions based on the current candidates, the confirmed evidence, and the remaining information gaps. Thus, the model autonomously determines the direction of search, while the Harness ensures both action validity and consistency with the current search state. We provide the detailed agent loop, tool definitions, and action constraints in Appendix~\ref{app:search_harness}.
\subsection{Harness-Grounded On-Policy Self-Distillation}
\label{train}
The harness introduced in Section~\ref{Stateful Search Harness} provides the decision interface on which we construct policy supervision. By maintaining cross-turn information in an explicit state and rendering a self-contained context for each invocation, it organizes a search trajectory into identifiable decision points connected by state transitions. At each point, the current evidence state provides the basis for constructing privileged feedback about candidate relevance, evidence coverage, and termination readiness. We use this feedback to condition a teacher policy that supervises the student within the same decision context and admissible action space. This aligns the granularity of supervision with that of execution and provides direct guidance for intermediate search behavior, reducing reliance on a single terminal reward.

Inspired by~\citet{zhao2026self}, we adopt On-Policy Self-Distillation (OPSD) to provide dense token-level training signals. H-Search structures multi-turn search as a sequence of stateless policy invocations connected through explicit search states. Each model invocation is treated as an independent inference process, in which the policy makes decisions solely on the basis of the current state. We construct privileged information to be process-oriented rather than outcome-oriented. Specifically, it indicates the extent to which the currently verified evidence covers the reference evidence, which evidence remains missing, whether it is safe to terminate the search at the current step, and which items in the evidence set are incorrect. Conditioned on the current state, we treat state-conditioned action generation as the unit of distillation and optimize the policy over complete search trajectories, rather than relying solely on answer-level correctness.
\paragraph{Privileged Policy Distillation.}
Specifically, the teacher and student are initialized from the same pretrained model, and supervision is derived from differences in the context available to each. The student receives only the standard search context provided by the Harness, while the frozen teacher additionally receives privileged information relevant to the current search state. For turn $t$, the two policies are represented as follows:

\begin{equation}
p_{t,\ell}
= \pi_{\theta}\left(\,\cdot \mid o_t, a_{t,<\ell}\right), \quad
q_{t,\ell}= \pi\left(\,\cdot \mid o_t, z_t, a_{t,<\ell}\right),
\end{equation}

here, $\theta$ denotes the student's trainable parameters, $\pi$ denotes the initial model, which remains frozen, $l$ indexes token positions in the action sequence, and $a_{t,<l}$ denotes the token prefix generated so far for the action at step $t$. The teacher thus provides supervision within the student's current decision context, without generating a separate search trajectory. After each action is executed, the Harness updates the state and constructs the context for the next step, ensuring that subsequent training states reflect the actual effects of the student's earlier decisions. We first initialize tool-use behavior within the search framework through a behavioral cloning warmup phase. During the formal training stage, for a sampled trajectory $\tau$, we calculate the forward KL distillation loss for each action step:

\begin{equation}
\mathcal{L}_{\mathrm{OPSD}}(\theta)
=
\frac{1}{|\mathcal{I}(\tau)|}
\sum_{t\in\mathcal{I}(\tau)}
w_t
\frac{1}{m_t}
\sum_{\ell=1}^{m_t}
D_{\mathrm{KL}}
\left(
q_{t,\ell}
\parallel
p_{t,\ell}
\right),
\end{equation}
where $\mathcal{I}(\tau)$ denotes the set of model decisions in the trajectory used for training, and $m_t$ is the number of tokens in the corresponding action. The step weights $w_t$ follow the SOD scheme proposed by \citet{zhong2026sod}, which adjusts the strength of supervision at different decision points based on teacher--student disagreement. Although the loss is computed separately for each step, training samples across steps remain connected through state transitions. Policy learning therefore spans the entire search process induced by the student's own actions. Please refer to the Appendix~\ref{TrainingDetail} for more detailed training information.

\section{Experiments}
\label{Experiments}
We organize our experiments around the following four research questions: 
\textbf{RQ1:} How does H-Search perform compared with existing retrieval and agentic search methods on open-domain benchmarks? \textbf{RQ2:} How does H-Search compare with search agents built on substantially larger backbone models? \textbf{RQ3:} To what extent can H-Search narrow the performance gap between smaller and larger search models? \textbf{RQ4:} How does the Harness-grounded on-policy self-distillation contribute to H-Search’s performance?
We first introduce the experimental settings, then address these questions through comparisons with existing methods, cross-model evaluations and component ablations, respectively.

\subsection{Experimental Settings}

\begin{table*}[h]
\centering
\caption{
Performance comparison on multi-hop and open-domain question answering
benchmarks. The \textcolor{bestblue}{\textbf{best}} results are highlighted in green, while the \textcolor{secondorange}{\textbf{second-best}} are highlighted in brown.
}
\fontsize{8.5}{10}\selectfont
\renewcommand{\arraystretch}{1.3}
\begin{adjustbox}{max width=0.98\linewidth}
\begin{tabular}{
llcccccc
@{\hspace{9pt}}
}
\toprule
\specialrule{\lightrulewidth}{2pt}{4pt}
\textbf{Method}
& \textbf{Backbone}
& \textbf{HotpotQA}
& \textbf{MuSiQue}
& \textbf{2Wiki}
& \textbf{NQ}
& \textbf{TriviaQA}
& \textbf{Avg.} \\
\midrule

\rowcolor{sectiongray}
\multicolumn{8}{l}{
\rule{0pt}{2ex}\textit{Training-free}
} \\
RAG
& Qwen3-Embedding-8B
& 0.318 & 0.275 & 0.273 & 0.643 & 0.589 & 0.420 \\

Hybrid Search
& BM25 + Qwen3-Embedding-8B
& 0.411 & 0.293 & 0.260 & 0.674 & 0.604 & 0.448 \\

IRCoT
& Qwen3.6-27B
& 0.678 & 0.432 & 0.597 & 0.745 & 0.652 & 0.621 \\

\midrule

\rowcolor{sectiongray}
\multicolumn{8}{l}{
\rule{0pt}{2ex}\textit{Search Agents}
} \\
Search-R1
& Qwen2.5-32B
& 0.482 & 0.381 & 0.477 & 0.735 & 0.597 & 0.534 \\

\arrayrulecolor{gray!55}
\midrule
\arrayrulecolor{black}

\multirow{1}{*}{CoRAG}
& Llama-3.1-8B-Inst.
& 0.643 & 0.540 & 0.638 & 0.597 & 0.454 & 0.574 \\

\arrayrulecolor{gray!55}
\midrule
\arrayrulecolor{black}

\multirow{2}{*}{FRUGAL}
& Qwen2.5-7B-Inst.
& 0.704 & 0.533 & 0.588 & 0.763 & 0.694 & 0.656 \\

& Qwen3.6-27B
& 0.729
& \textcolor{secondorange}{\textbf{0.547}} & 0.612 & \textcolor{secondorange}{\textbf{0.817}}
& 0.725 & 0.686 \\

\midrule

\rowcolor{sectiongray}
\multicolumn{8}{l}{
\rule{0pt}{2ex}\textit{Ours}
} \\
H-Search-Lite
& Qwen3.6-27B & \textcolor{secondorange}{\textbf{0.756}} & 0.513
& \textcolor{secondorange}{\textbf{0.759}} & 0.800 & \textcolor{secondorange}{\textbf{0.739}} & \textcolor{secondorange}{\textbf{0.714}} \\

H-Search-OPSD
& Qwen3.6-27B
& \textcolor{bestblue}{\textbf{0.839}}
& \textcolor{bestblue}{\textbf{0.657}}
& \textcolor{bestblue}{\textbf{0.882}}
& \textcolor{bestblue}{\textbf{0.820}}
& \textcolor{bestblue}{\textbf{0.792}}
& \textcolor{bestblue}{\textbf{0.798}} \\

\specialrule{\lightrulewidth}{4pt}{2pt}
\bottomrule
\end{tabular}
\end{adjustbox}
\label{t1}
\end{table*}

\paragraph{Benchmarks and Evaluation.}
We evaluate H-Search on five widely used knowledge-intensive question answering benchmarks: HotpotQA \citep{yang2018hotpotqa}, MuSiQue \citep{trivedi2022musique}, 2WikiMultiHopQA(2Wiki) \citep{ho2020constructing}, Natural Questions(NQ) \citep{kwiatkowski2019natural}, and TriviaQA\citep{joshi2017triviaqa}. All five benchmarks are evaluated in an open-domain retrieval setting, where agents search an external corpus without being provided with pre-selected supporting passages. HotpotQA, MuSiQue, and 2Wiki assess multi-hop information seeking that requires gathering evidence across multiple passages, while NQ and TriviaQA cover naturally occurring search queries and diverse factual questions. We evaluate retrieval recall on the final selected evidence set returned by the agent, computed for each query as the fraction of gold-standard supporting documents retrieved by the agent and then averaged across queries.

\paragraph{Baselines.} We compare H-Search with three training-free baselines, including standard RAG using Qwen3-Embedding-8B~\citep{qwen3embedding} as the embedding model, Hybrid Search, which combines dense vector retrieval with BM25-based retrieval, and IRCoT~\citep{trivedi2023interleaving}, and three representative trained search agents, namely Search-R1~\citep{jin2025search}, CoRAG ~\citep{wang2026chain}, and FrugalRAG~\citep{java2026frugalrag}. For all methods, we use the same retriever and set the top-$k$ parameter to 10 when calculating recall. 
Since the final documents in H-Search are selected by the Harness, our method does not enforce a fixed top-$k$ value; however, we set an upper limit of 10 for the final document pool.
We also trained and evaluated FrugalRAG using the same Qwen3.6-27B backbone network as H-search.~\citep{qwen3.6-27b}. Further details regarding the baselines can be found in the Appendix\ref{app:baselines}.

\paragraph{Implementation.} 
We construct the training set by sampling 1,000 examples from the training split of each of the five benchmarks, yielding 5,000 examples in total. All benchmarks use a full-Wikipedia retrieval setting. We use Qwen3.6-27B as the policy backbone and initialize the student and frozen teacher from the same pretrained checkpoint. Training follows the harness-grounded on-policy self-distillation objective in Section \ref{train}. At inference, neither the teacher nor privileged information is used. We evaluate the same OPSD-trained checkpoint under two inference configurations. H-Search-OPSD enables thinking and permits up to 20 decision rounds, whereas H-Search-Lite disables thinking and permits up to 8 decision rounds. Both configurations use the same trained model parameters and search harness. The OPSD ablation compares the initial and trained checkpoints under the same Lite inference configuration.

\subsection{Main Results (RQ1)}
\paragraph{H-Search achieves advanced retrieval performance.} Table \ref{t1} shows that H-Search-OPSD obtains 79.8\% average recall, exceeding FrugalRAG with Qwen3.6-27B, the strongest baseline by average score, by 11.2 percentage points. It also outperforms the CoRAG and IRCoT configurations by 22.4 and 17.7 points, respectively. H-Search-OPSD ranks first on all five benchmarks, indicating that its overall advantage extends across both multi-hop and factual question datasets. The gains are most pronounced on multi-hop benchmarks. Relative to FrugalRAG with the same backbone, H-Search-OPSD improves recall by 11.0 points on HotpotQA, 11.0 on MuSiQue, and 27.0 on 2Wiki. Averaged over these three datasets, recall increases from 62.9\% to 79.3\%. The method also improves TriviaQA from 72.5\% to 79.2\%, while NQ changes only modestly, from 81.7\% to 82.0\%. Thus, the largest contribution to the average improvement comes from answer-bearing document retrieval on multi-hop questions. This pattern is consistent with the intended use of H-Search for search tasks involving dependencies across multiple decisions.

\subsection{Larger-Model Comparison (RQ2)}
\paragraph{Search-R1 Baselines.} In Table \ref{tab:frontier-models}, the strongest larger-backbone Search-R1 configuration, GLM5.2, achieves 72\% average recall, compared with 79.8\% for H-Search-OPSD. The resulting advantage is 7.8pp. H-Search-OPSD also exceeds the GPT-OSS-120B, KIMI-K3, and Qwen3.5-397B configurations by 9.3, 13.1, and 19.2 points, respectively. Across individual benchmarks, the 27B system exceeds every Search-R1 configuration on HotpotQA, MuSiQue, 2Wiki, and TriviaQA, and matches the best Search-R1 result on NQ at the reported precision.

\paragraph{H-Search Baselines.} H-Search-OPSD outperforms GPT-OSS-120B with H-Search on every benchmark and ranks second overall in Table 2, behind GLM5.2 with H-Search. These results show that substantially larger backbones do not necessarily produce stronger retrieval systems: the trained 27B policy, together with its execution framework, exceeds several larger-model configurations.

\subsection{Bridging the Model-Scale Gap (RQ3)}
\begin{wrapfigure}{l}{0.45\textwidth}
    \centering
    \includegraphics[width=\linewidth]{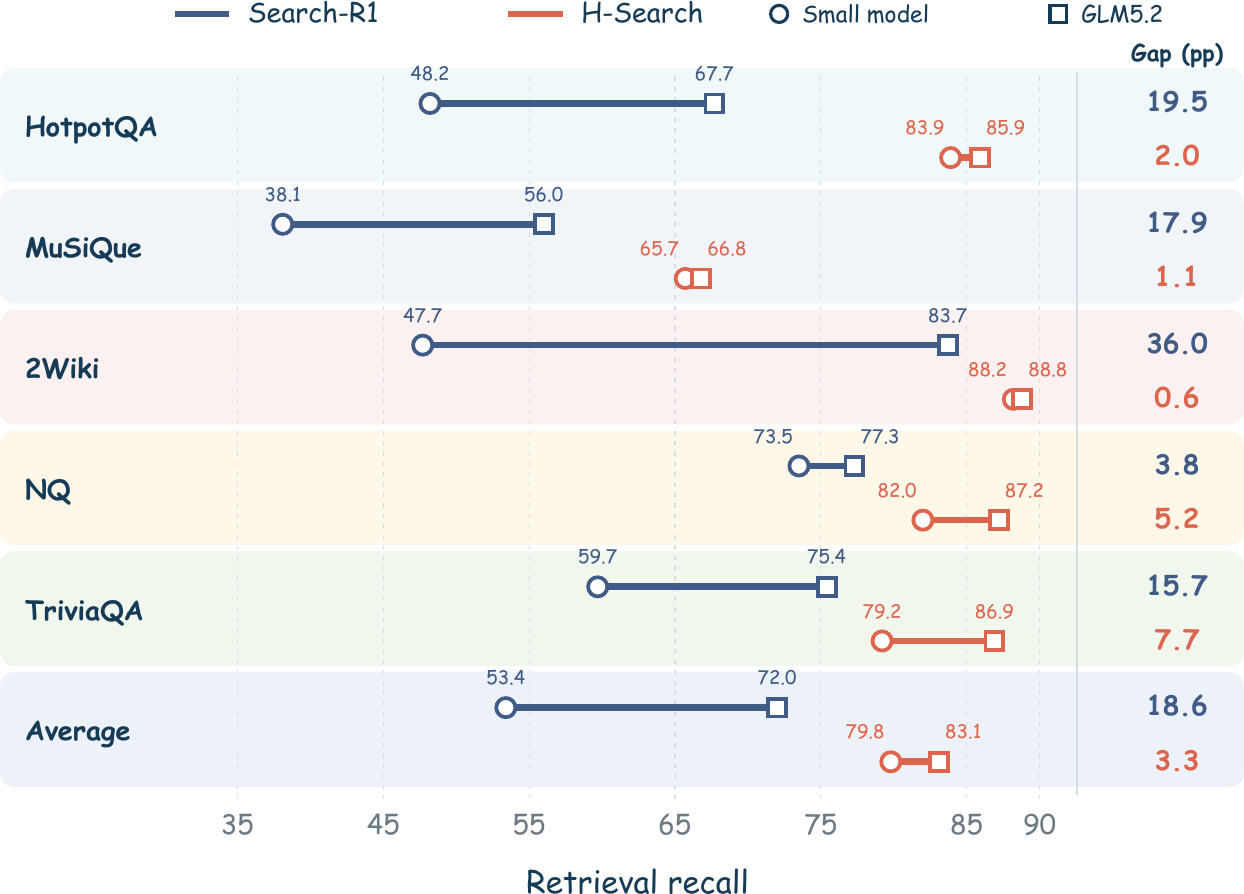}
    \caption{
         Retrieval performance across model scales.
    }
    \label{fig:model-scale-gap}
\end{wrapfigure}
We examine how closely a smaller model trained with H-Search
can approach the retrieval performance of a substantially
larger backbone. Figure~\ref{fig:model-scale-gap} compares Qwen2.5-32B and GLM5.2 under Search-R1, alongside Qwen3.6-27B trained with
OPSD and GLM5.2 under H-Search. The GLM5.2-based H-Search system uses the harness without additional training.
H-Search exhibits a substantially smaller performance gap
between the evaluated small- and large-model configurations.
Under Search-R1, average recall increases from 53.4\% with
Qwen2.5-32B to 72.0\% with GLM5.2, a difference of
18.6 percentage points. In contrast, H-Search-OPSD with Qwen3.6-27B achieves 79.8\%, only 3.3 points below GLM5.2 with H-Search (83.1\%). The trained 27B system also exceeds GLM5.2 with Search-R1
by 7.8 points, showing that competitive retrieval performance
can be achieved without relying on the largest backbone.
The gap is particularly small on multi-hop benchmarks.
Under H-Search, the differences are only 2.0, 1.1, and
0.6 percentage points on HotpotQA, MuSiQue, and 2Wiki,
respectively, compared with 19.5, 17.9, and 36.0 points
under Search-R1. On NQ and TriviaQA, the trained 27B model also approaches the GLM5.2-based H-Search system, with gaps of 5.2 and
7.7 points, respectively. Together, these results suggest that combining explicit search-state management with OPSD training can bring a smaller model close to a substantially larger-model
search system. This comparison reflects the combined effects of the
backbone, harness, and policy training.

\begin{table*}[h]
\centering
\caption{Retrieval recall performance of large-parameter backbone models using different retrieval frameworks and our model across five open-domain benchmarks. The \textcolor{bestblue}{\textbf{best}} results are highlighted
in green, while the \textcolor{secondorange}{\textbf{second-best}} results
are highlighted in brown.
}
\fontsize{8.5}{10}\selectfont
\renewcommand{\arraystretch}{1.3}
\begin{adjustbox}{max width=0.98\linewidth}
\begin{tabular}{
llcccccc
@{\hspace{9pt}}
}
\toprule
\specialrule{\lightrulewidth}{2pt}{4pt}
\textbf{Method}
& \textbf{Backbone}
& \textbf{HotpotQA}
& \textbf{MuSiQue}
& \textbf{2Wiki}
& \textbf{NQ}
& \textbf{TriviaQA}
& \textbf{Avg.} \\
\midrule

\rowcolor{sectiongray}
\multicolumn{8}{l}{
\rule{0pt}{2ex}\textit{Frontier Models (Search-R1)}
} \\

\multirow{4}{*}{Search-R1}
& Qwen3.5-397B
& 0.533
& 0.437
& 0.585
& 0.753
& 0.723
& 0.606 \\

& GPT-OSS-120B
& 0.690
& 0.507
& 0.798
& \textcolor{secondorange}{\textbf{0.820}}
& 0.708
& 0.705 \\

& GLM5.2
& 0.677 & 0.560 & 0.837 & 0.773 & 0.754 & 0.720 \\

& KIMI-K3
& 0.593 & 0.497 & 0.717 & 0.767 & 0.763 & 0.667 \\

\midrule

\rowcolor{sectiongray}
\multicolumn{8}{l}{
\rule{0pt}{2ex}\textit{Frontier Models (Ours)}
} \\

\multirow{2}{*}{H-Search}
& GPT-OSS-120B
& 0.721
& 0.524
& 0.809
& 0.811
& 0.714
& 0.716 \\

& GLM5.2
& \textcolor{bestblue}{\textbf{0.859}}
& \textcolor{bestblue}{\textbf{0.668}}
& \textcolor{bestblue}{\textbf{0.888}}
& \textcolor{bestblue}{\textbf{0.872}}
& \textcolor{bestblue}{\textbf{0.869}}
& \textcolor{bestblue}{\textbf{0.831}} \\

\midrule

\rowcolor{sectiongray}
\multicolumn{8}{l}{
\rule{0pt}{2ex}\textit{Ours}
} \\

H-Search-OPSD
& Qwen3.6-27B
& \textcolor{secondorange}{\textbf{0.839}}
& \textcolor{secondorange}{\textbf{0.657}}
& \textcolor{secondorange}{\textbf{0.882}}
& \textcolor{secondorange}{\textbf{0.820}}
& \textcolor{secondorange}{\textbf{0.792}}
& \textcolor{secondorange}{\textbf{0.798}} \\

\specialrule{\lightrulewidth}{4pt}{2pt}
\bottomrule
\end{tabular}
\end{adjustbox}
\label{tab:frontier-models}
\end{table*}

\subsection{Component Analysis (RQ4)}
\label{rq4}

\begin{wrapfigure}{r}{0.45\textwidth}
    \centering
    \includegraphics[width=\linewidth]{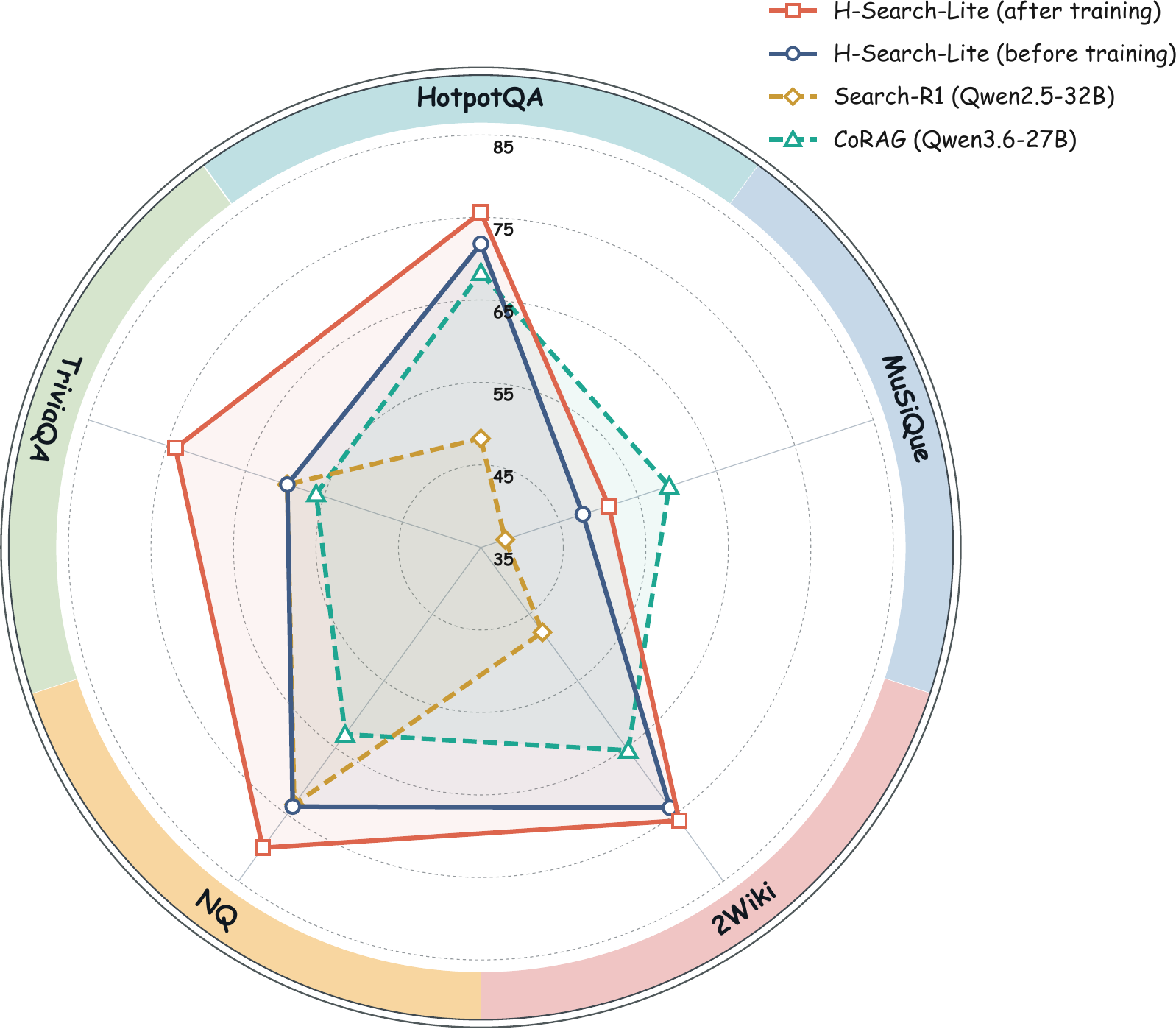}
    \caption{
        Retrieval recall across five benchmarks.
        H-Search-Lite before and after training is compared with
        Search-R1 and CoRAG. The radial axis spans 35\%--85\%.
    }
    \label{fig:training-recall}
\end{wrapfigure}

\paragraph{Training Ablation.} To assess the contribution of harness-grounded OPSD, we compare H-Search-Lite before and after training.
As shown in Figure~\ref{fig:training-recall}, OPSD improves
document-level recall across all five benchmarks, increasing
the macro-average from 65.46\% to 71.35\% (+5.90 percentage points).
The largest improvement occurs on TriviaQA, where recall rises
from 59.67\% to 73.94\% (+14.27 points), followed by NQ
with a gain of 6.18 points. Improvements also extend to the three multi-hop benchmarks: 3.80 points on HotpotQA, 3.30 on MuSiQue, and 1.93 on 2Wiki. These results indicate that the benefits of OPSD are not restricted to multi-hop retrieval; substantial gains also
arise on factual question benchmarks. The trained policy outperforms Search-R1 on all five datasets and CoRAG on four, with MuSiQue remaining an exception. Overall, the before-and-after comparison supports the effectiveness of harness-grounded OPSD in improving evidence coverage within H-Search-Lite. Please refer to the Appendix~\ref{addab} for further details on the ablation.

\section{Conclusion}

We presented H-Search, a framework that couples explicit search-state management with decision-level policy training for agentic search. Its stateful harness organizes multi-turn interactions into
self-contained policy invocations, while harness-grounded on-policy
self-distillation uses privileged evidence-state information to
supervise decisions along student-generated trajectories.
The resulting policy operates within the same harness at inference,
without privileged information or teacher access.
Across five open-domain retrieval benchmarks, H-Search with a 27B
backbone achieves 79.8\% average document-level recall, exceeding
the strongest evaluated same-backbone baseline by 11.2 percentage
points and outperforming several substantially larger-model
configurations.
The ablation study further shows consistent improvements across
all five benchmarks for H-Search-Lite.
Together, these findings support treating the search harness
as an integral part of policy learning and highlight the value
of aligning decision-level supervision with explicit evidence states. Future work will explore this approach in open-web environments and examine how improved evidence coverage translates into downstream answer quality.

\subsection*{AI Use Statement}
We used large language models to assist with language editing and
code development during the preparation of this work. Specifically,
these tools helped improve the clarity and grammar of the manuscript
and supported code implementation and debugging. All AI-assisted
content incorporated into the paper or codebase was reviewed and
verified by the authors. The authors take full responsibility for
the accuracy and integrity of the paper, including its methodology,
experimental results, interpretations, and conclusions.

\subsection*{Reproducibility statement}

To facilitate reproducibility, we provide the source code for our
method in the supplementary materials. The main paper describes
the proposed method and experimental setup, while the appendix
provides additional implementation details and hyperparameter
settings. Together, these materials support the reproduction
of our experiments and verification of the reported results.

\clearpage

\appendix

\section{Baseline Methods}
\label{app:baselines}

We compare H-Search with six baselines covering single-step retrieval, iterative reasoning-guided retrieval, and learned search policies. All baselines utilized the agent architectures and toolsets specified in their original papers, while employing consistent top-$k$ values and retrievers.
\paragraph{RAG.}
A single-step dense retrieval baseline using Qwen3-Embedding-8B to retrieve documents relevant to the original question.

\paragraph{Hybrid Search.}
A retrieval baseline combining dense semantic matching with BM25-based lexical retrieval.

\paragraph{IRCoT.}
IRCoT \citep{trivedi2023interleaving} interleaves chain-of-thought reasoning with retrieval, using intermediate reasoning steps to guide subsequent searches. We use Qwen3.6-27B as its backbone.

\paragraph{Search-R1.}
Search-R1 \citep{jin2025search} trains language models to interleave reasoning and search through reinforcement learning with outcome-based rewards. We evaluate its Qwen2.5-32B configuration and additional larger-backbone configurations.

\paragraph{CoRAG.}
CoRAG \citep{wang2026chain} learns from retrieval chains constructed through rejection sampling, enabling iterative query reformulation and evidence gathering. We evaluate Llama-3.1-8B-Instruct and Qwen3.6-27B configurations.

\paragraph{FrugalRAG.}
FrugalRAG \citep{java2026frugalrag} combines supervised exploration training with reinforcement learning to balance answer correctness and retrieval cost. We evaluate Qwen2.5-7B-Instruct and Qwen3.6-27B configurations.

\section{Additional Results}

\subsection{Additional Training Performance}
\label{addab}
We further compare H-Search, H-Search-GRPO, and H-Search-OPSD
to examine the effects of policy training within the H-Search
framework. Table~\ref{addablation} reports document-level recall on the five
benchmarks, complementing the H-Search-Lite training ablation
in Section~\ref{rq4}.

\paragraph{Overall performance.}
H-Search-OPSD achieves the highest macro-average recall of
79.8\%, compared with 76.3\% for H-Search and 77.6\% for
H-Search-GRPO.  GRPO improves average recall by 1.3 points, although its gains are less consistent: recall increases on four
benchmarks but decreases slightly on 2Wiki, from 86.6\% to
86.4\%. We also isolated the effects of the warmup phase and OPSD during training, observing that the full OPSD implementation led to significant performance gains.

\paragraph{Comparison across benchmarks.}
OPSD outperformed GRPO across all five benchmarks.
On the multi-hop datasets, it improves recall by 1.2 percentage
points on HotpotQA, 1.2 on MuSiQue, and 1.8 on 2Wiki.
Meanwhile, it achieved gains of 2.8 and 4.1 percentage points on NQ and TriviaQA, respectively.
These results support the effectiveness of harness-grounded OPSD for improving evidence coverage, with a consistent advantage over GRPO across the three multi-hop benchmarks. This pattern is consistent with the motivation for providing decision-level guidance aligned with the evolving evidence state.

\paragraph{Interaction with inference configuration.}
The improvement from OPSD is smaller under the full H-Search
configuration than under H-Search-Lite: macro-average recall
increases by 3.5 percentage points in Table~\ref{addablation}, compared with approximately 5.9 points in the Lite ablation.
One possible explanation is that the full configuration
enables thinking and allows up to 20 decision rounds,
whereas Lite disables thinking and permits only 8 rounds.
Deeper reasoning and additional opportunities to search
may help the initial policy recover from intermediate
errors, leaving less room for improvement through training.
Conversely, under the more constrained Lite configuration,
better decisions at each step may have a larger effect
on final evidence coverage. 

\begin{table}[ht]
\centering
\caption{Recall of H-Search and its trained variants across
five retrieval benchmarks. Best and second-best results are
highlighted in blue and orange, respectively.}
\label{addablation}

\fontsize{8.5}{10}\selectfont
\renewcommand{\arraystretch}{1.3}
\setlength{\tabcolsep}{5pt}

\begin{adjustbox}{max width=0.98\linewidth}
\begin{tabular}{lccccc@{\hspace{9pt}}}
\toprule
\specialrule{\lightrulewidth}{2pt}{4pt}
\textbf{Method}
& \textbf{HotpotQA}
& \textbf{MuSiQue}
& \textbf{2Wiki}
& \textbf{NQ}
& \textbf{TriviaQA} \\
\midrule

\rowcolor{sectiongray}
\multicolumn{6}{l}{
\rule{0pt}{2ex}\textit{Without Training}
} \\

H-Search
& 0.814 & 0.638 & 0.866 & 0.769 & 0.728 \\

\midrule

\rowcolor{sectiongray}
\multicolumn{6}{l}{
\rule{0pt}{2ex}\textit{With Training}
} \\

H-Search-BC-only
& 0.821
& 0.644
& \textcolor{secondorange}{\textbf{0.871}}
& 0.783
& 0.745 \\

H-Search-GRPO
& \textcolor{secondorange}{\textbf{0.827}}
& \textcolor{secondorange}{\textbf{0.645}}
& 0.864
& \textcolor{secondorange}{\textbf{0.792}}
& \textcolor{secondorange}{\textbf{0.751}} \\

H-Search-OPSD
& \textcolor{bestblue}{\textbf{0.839}}
& \textcolor{bestblue}{\textbf{0.657}}
& \textcolor{bestblue}{\textbf{0.882}}
& \textcolor{bestblue}{\textbf{0.820}}
& \textcolor{bestblue}{\textbf{0.792}} \\

\specialrule{\lightrulewidth}{4pt}{2pt}
\bottomrule
\end{tabular}
\end{adjustbox}
\end{table}

\subsection{Additional Ablation}

\paragraph{Ablation of harness components.}
To examine the contributions of individual harness components,
we compare H-Search with two variants that remove structured state
or state-dependent action constraints.
Table~\ref{addablation3} shows that removing
either component reduces document-level recall on all five benchmarks.
Removing structured state causes the larger degradation, reducing
macro-average recall from 76.3\% to 46.7\%
(a drop of 29.6 percentage points).
The decline is particularly pronounced on multi-hop benchmarks:
recall drops by 37.7 points on HotpotQA, 31.2 points on MuSiQue,
and 57.1 points on 2Wiki, compared with 10.9 points on NQ
and 11.1 points on TriviaQA. These results suggest that explicit
state organization is especially important for search tasks requiring
evidence accumulation across multiple hops.

Removing state-dependent action constraints leads to a smaller
but consistent decline, lowering macro-average recall from 76.3\%
to 74.0\% (a drop of 2.3 percentage points).
The per-dataset reductions range from 1.2 to 3.7 points,
with the largest decrease on 2Wiki.
Together, these results support the contributions of both
structured state and state-dependent action constraints
to the retrieval performance of H-Search.

\begin{table}[ht]
\centering
\caption{Ablation of structured state and state-dependent action constraints in H-Search. We report document-level recall on five retrieval benchmarks. Each ablated variant removes the indicated component. The \textcolor{bestblue}{\textbf{best}} results are highlighted in blue.}
\label{addablation3}
\fontsize{8.5}{10}\selectfont
\renewcommand{\arraystretch}{1.3}
\setlength{\tabcolsep}{5pt}

\begin{adjustbox}{max width=0.98\linewidth}
\begin{tabular}{lccccc@{\hspace{9pt}}}
\toprule
\specialrule{\lightrulewidth}{2pt}{4pt}
\textbf{Method}
& \textbf{HotpotQA}
& \textbf{MuSiQue}
& \textbf{2Wiki}
& \textbf{NQ}
& \textbf{TriviaQA} \\
\midrule

H-Search$_{w/o \text{structured state}}$
& 0.437 & 0.326 & 0.295 & 0.660 & 0.617 \\

H-Search$_{w/o \text{action constraints}}$
& 0.802
& 0.617
& 0.829
& 0.754
& 0.696 \\

H-Search
& \textcolor{bestblue}{\textbf{0.814}}
& \textcolor{bestblue}{\textbf{0.638}}
& \textcolor{bestblue}{\textbf{0.866}}
& \textcolor{bestblue}{\textbf{0.769}}
& \textcolor{bestblue}{\textbf{0.728}} \\

\specialrule{\lightrulewidth}{4pt}{2pt}
\bottomrule
\end{tabular}
\end{adjustbox}
\end{table}

\subsection{Effect of the Search Harness}

To evaluate the contribution of the search harness, we compare
different search architectures using the same Qwen3.6-27B checkpoint
without additional task-specific training.
We instantiate the search workflows of CoRAG and FrugalRAG with this
shared backbone and compare them with H-Search before policy training.
As shown in Table~\ref{addablation2}, H-Search achieves the
highest document-level recall on all five benchmarks.
Its macro-average recall reaches 76.3\%, compared with 62.3\% for
CoRAG and 64.7\% for FrugalRAG, corresponding to improvements of
14.0 and 11.6 percentage points, respectively.
Relative to the stronger baseline on each dataset, H-Search improves
recall by 12.3 points on HotpotQA, 4.8 on MuSiQue, 21.2 on 2Wiki,
8.2 on NQ, and 1.0 on TriviaQA.
These results show that the H-Search harness provides substantial
benefits even without policy training.

\begin{table}[ht]
\centering
\caption{Effect of the search harness on  recall across five
retrieval benchmarks. All methods use the same Qwen3.6-27B checkpoint without additional task-specific training. }
\label{addablation2}

\fontsize{8.5}{10}\selectfont
\renewcommand{\arraystretch}{1.3}
\setlength{\tabcolsep}{5pt}

\begin{adjustbox}{max width=0.98\linewidth}
\begin{tabular}{llccccc@{\hspace{9pt}}}
\toprule
\specialrule{\lightrulewidth}{2pt}{4pt}
\textbf{Method}
& \textbf{Backbone}
& \textbf{HotpotQA}
& \textbf{MuSiQue}
& \textbf{2Wiki}
& \textbf{NQ}
& \textbf{TriviaQA} \\
\midrule

CoRAG
 & Qwen3.6-27B
 & 0.683
 & 0.590
 & 0.654
 & 0.630
 & 0.560
\\
FRUGALRAG
 & Qwen3.6-27B
 & 0.691
 & 0.529
 & 0.609
 & 0.687
 & 0.718
\\

H-Search
 & Qwen3.6-27B
& \textcolor{bestblue}{\textbf{0.814 }}
& \textcolor{bestblue}{\textbf{0.638}}
& \textcolor{bestblue}{\textbf{0.866}}
& \textcolor{bestblue}{\textbf{0.769}}
& \textcolor{bestblue}{\textbf{0.728}} \\

\specialrule{\lightrulewidth}{4pt}{2pt}
\bottomrule
\end{tabular}
\end{adjustbox}
\end{table}

\subsection{Harness-Training Interaction}

To examine whether the benefits of OPSD depend on the search framework, we apply it to both FrugalRAG and H-Search, consistently using the Qwen3.6-27B model. As shown in Table~\ref{addablation5}, OPSD improves macro-average recall by 0.84 percentage points under FrugalRAG, compared with 3.50 points under H-Search.
The training gain is therefore 2.66 points larger under H-Search, despite its higher initial recall. Moreover, H-Search improves on all five benchmarks, whereas
FrugalRAG shows smaller gains on four benchmarks and a slight decline on 2Wiki.  They support the complementarity of the H-Search harness and policy training, consistent with our design of using explicit evidence states and state-conditioned decisions as a shared interface for execution and supervision.

\begin{table}[ht]
\centering
\caption{Recall with and without OPSD under the FrugalRAG and H-Search frameworks across five retrieval benchmarks. Avg.\ denotes the unweighted mean across benchmarks, and $\Delta$ Avg.\ reports the absolute improvement from training
within each framework.}
\label{addablation5}

\fontsize{8.5}{10}\selectfont
\renewcommand{\arraystretch}{1.3}
\setlength{\tabcolsep}{5pt}

\begin{adjustbox}{max width=0.98\linewidth}
\begin{tabular}{lccccccc@{\hspace{9pt}}}
\toprule
\specialrule{\lightrulewidth}{2pt}{4pt}
\textbf{Method}
& \textbf{HotpotQA}
& \textbf{MuSiQue}
& \textbf{2Wiki}
& \textbf{NQ}
& \textbf{TriviaQA}
& \textbf{Avg.}
& \textbf{$\Delta$ Avg.} \\
\midrule

\rowcolor{sectiongray}
\multicolumn{8}{l}{
\rule{0pt}{2ex}\textit{FRUGALRAG}
} \\

FRUGAL
& 0.691
& 0.529
& 0.609
& 0.687
& 0.718
& 0.647
& \multirow{2}{*}{+0.84 pp} \\

FRUGAL-OPSD
& 0.704
& 0.538
& 0.603
& 0.702
& 0.729
& 0.655
& \\

\midrule
\rowcolor{sectiongray}
\multicolumn{8}{l}{
\rule{0pt}{2ex}\textit{H-Search (Ours)}
} \\

H-Search
& 0.814
& 0.638
& 0.866
& 0.769
& 0.728
& 0.763
& \multirow{2}{*}{+3.50 pp} \\

H-Search-OPSD
& 0.839
& 0.657
& 0.882
& 0.820
& 0.792
& 0.798
& \\

\specialrule{\lightrulewidth}{4pt}{2pt}
\bottomrule
\end{tabular}
\end{adjustbox}
\end{table}

\subsection{Noise in Selected Evidence}
\label{Precision}
To examine whether H-Search reduces noise in the final selected
evidence, we evaluate document-level precision on three multi-hop
benchmarks. Precision reflects the proportion of gold supporting documents
in the final selected evidence set; higher precision therefore
indicates less noise in the selected evidence.

\begin{table}[h]
\centering
\caption{Document-level recall (R), precision (P), and F1 on
three multi-hop benchmarks. F1 is computed as the harmonic mean
of dataset-level R and P. Avg.\ denotes the unweighted mean
across benchmarks. \textcolor{bestblue}{\textbf{Best}}  and \textcolor{secondorange}{\textbf{second-best}} results are highlighted
in blue and orange, respectively.}
\label{tab:evidence_precision}

\fontsize{8.5}{10}\selectfont
\renewcommand{\arraystretch}{1.3}
\setlength{\tabcolsep}{3pt}

\begin{adjustbox}{max width=0.98\linewidth}
\begin{tabular}{l*{12}{c}@{\hspace{9pt}}}
\toprule
\specialrule{\lightrulewidth}{2pt}{4pt}
& \multicolumn{3}{c}{\textbf{HotpotQA}}
& \multicolumn{3}{c}{\textbf{MuSiQue}}
& \multicolumn{3}{c}{\textbf{2Wiki}}
& \multicolumn{3}{c}{\textbf{Avg.}} \\
\cmidrule(lr){2-4}
\cmidrule(lr){5-7}
\cmidrule(lr){8-10}
\cmidrule(lr){11-13}
\textbf{Method}
& \textbf{R} & \textbf{P} & \textbf{F1}
& \textbf{R} & \textbf{P} & \textbf{F1}
& \textbf{R} & \textbf{P} & \textbf{F1}
& \textbf{R} & \textbf{P} & \textbf{F1} \\
\midrule

\rowcolor{sectiongray}
\multicolumn{13}{l}{
\rule{0pt}{2ex}\textit{Baselines}
} \\

IRCoT
& 0.678 & 0.145 & 0.239
& 0.432 & 0.137 & 0.208
& 0.597 & 0.193 & 0.292
& 0.569 & 0.158 & 0.246 \\

Search-R1
& 0.482 & 0.167 & 0.248
& 0.381 & 0.140 & 0.205
& 0.477 & 0.212 & 0.294
& 0.447 & 0.173 & 0.249 \\

CoRAG
& 0.643 & 0.264 & 0.374
& \textcolor{secondorange}{\textbf{0.540}}
& 0.137 & 0.219
& \textcolor{secondorange}{\textbf{0.638}}
& 0.315 & 0.422
& 0.607 & 0.239 & 0.338 \\

FrugalRAG
& \textcolor{secondorange}{\textbf{0.704}}
& \textcolor{secondorange}{\textbf{0.374}}
& \textcolor{secondorange}{\textbf{0.488}}
& 0.533
& \textcolor{secondorange}{\textbf{0.247}}
& \textcolor{secondorange}{\textbf{0.338}}
& 0.588
& \textcolor{secondorange}{\textbf{0.425}}
& \textcolor{secondorange}{\textbf{0.493}}
& \textcolor{secondorange}{\textbf{0.608}}
& \textcolor{secondorange}{\textbf{0.349}}
& \textcolor{secondorange}{\textbf{0.440}} \\

\midrule

\rowcolor{sectiongray}
\multicolumn{13}{l}{
\rule{0pt}{2ex}\textit{Ours}
} \\

H-Search-OPSD
& \textcolor{bestblue}{\textbf{0.839}}
& \textcolor{bestblue}{\textbf{0.485}}
& \textcolor{bestblue}{\textbf{0.615}}
& \textcolor{bestblue}{\textbf{0.657}}
& \textcolor{bestblue}{\textbf{0.311}}
& \textcolor{bestblue}{\textbf{0.422}}
& \textcolor{bestblue}{\textbf{0.882}}
& \textcolor{bestblue}{\textbf{0.652}}
& \textcolor{bestblue}{\textbf{0.750}}
& \textcolor{bestblue}{\textbf{0.793}}
& \textcolor{bestblue}{\textbf{0.483}}
& \textcolor{bestblue}{\textbf{0.596}} \\

\specialrule{\lightrulewidth}{4pt}{2pt}
\bottomrule
\end{tabular}
\end{adjustbox}
\end{table}

As shown in Table~\ref{tab:evidence_precision}, H-Search-OPSD
achieves the highest precision on all three benchmarks.
Compared with FrugalRAG, the strongest baseline in this comparison,
it reduces the average noise ratio from 65.1\% to 51.7\%,
a decrease of 13.4 percentage points.
The reductions are 11.1 points on HotpotQA, 6.4 on MuSiQue,
and 22.7 on 2Wiki.
Meanwhile, average recall improves from 60.8\% to 79.3\%,
showing that H-Search produces less noisy evidence sets
while improving gold evidence coverage.

\section{Search Harness and Training Details}

\subsection{Search Harness}
\label{app:search_harness}

H-Search implements a render--decide--execute--update loop.
The harness maintains persistent search state, constructs
the context for each policy invocation, and restricts the
available actions according to the current control phase.
Table~\ref{tab:harness_tools} summarizes its tools.

The persistent state $s_t$ maintains the query plan, candidate pool, selected evidence, available document content, search history, and control phase. Retrieval adds candidates, reading records their content and reading status, and curation moves selected candidates into the evidence set. The selected evidence set hold at most 10 documents. When the candidate pool overflows, candidates are retained according to retrieval frequency and accumulated ranking
scores. 

\paragraph{Self-contained policy invocations.}
Before each decision, the harness constructs
$o_t = R(s_t)$ from the current state.
For ordinary search decisions, the context includes the
query plan, selected document identifiers, ranked document
excerpts, reading status, and previous search attempts.
A bounded window of recent interactions supplies local
feedback without replaying the complete trajectory.
For sufficiency decisions, the context instead emphasizes
the selected evidence and remaining information needs.
The harness also supplies the tool schemas available in
the current phase.
Each policy invocation therefore receives its cross-turn
information through an explicitly reconstructed input,
without requiring the model to retain internal state
across calls.
Decisions remain connected through the persistent state:
the executed action and its tool feedback update $s_t$,
from which the next decision context is constructed.

\begin{table}[]
    \centering
    \small
    \renewcommand{\arraystretch}{1.35}
    \setlength{\tabcolsep}{6pt}
    \setlength{\fboxsep}{2.5pt}
    \caption{Tools provided by the H-Search harness and their
    functions. Tool availability is governed by the current
    search state.}
    \label{tab:harness_tools}
    \begin{tabular}{@{}p{0.35\linewidth}p{0.60\linewidth}@{}}
        \toprule
        \textbf{Tool} & \textbf{Function} \\
        \midrule
        \colorbox[HTML]{FFF2DF}{%
            \textcolor[HTML]{996024}{%
                \strut\ttfamily\bfseries Query planning}}
        & Decompose the question into entities and retrieval
          clues, and construct focused subqueries. \\
        \addlinespace[4pt]

        \colorbox[HTML]{E8EFFA}{%
            \textcolor[HTML]{365D96}{%
                \strut\ttfamily\bfseries Search}}
        & Retrieve candidate results for the query. \\
        \addlinespace[4pt]

        \colorbox[HTML]{E5F3F7}{%
            \textcolor[HTML]{2C7185}{%
                \strut\ttfamily\bfseries Grep}}
        & Retrieve candidates using keywords or entity names. \\
        \addlinespace[4pt]

        \colorbox[HTML]{F7EBDD}{%
            \textcolor[HTML]{8C6540}{%
                \strut\ttfamily\bfseries Read}}
        & Read the available content of a specified document
          and update its reading status. \\
        \addlinespace[4pt]

        \colorbox[HTML]{EDF4E3}{%
            \textcolor[HTML]{527B39}{%
                \strut\ttfamily\bfseries Select}}
        & Add selected candidates to the persistent evidence
          set, optionally with supporting notes. \\
        \addlinespace[4pt]

        \colorbox[HTML]{F8E8E6}{%
            \textcolor[HTML]{A0524D}{%
                \strut\ttfamily\bfseries Judge sufficiency}}
        & Assess evidence sufficiency and request termination
          or identify a concrete follow-up target. \\
        \addlinespace[4pt]

        \colorbox[HTML]{E2F2EC}{%
            \textcolor[HTML]{347A63}{%
                \strut\ttfamily\bfseries Finish}}
        & End the search and return the selected evidence set. \\
        \bottomrule
    \end{tabular}
\end{table}

\paragraph{State-dependent action constraints.}
The harness computes $A_t = \Gamma(s_t)$ and exposes only
the corresponding tool schemas to the policy.
After retrieval produces candidates, the available actions
are restricted to Read and Select until curation completes
the current retrieval cycle.
A successful curation with nonempty evidence makes
Judge sufficiency the only action available at the next
policy invocation.
The same policy then requests termination or identifies
a concrete missing fact or follow-up query.
The controller executes another search when an actionable
continuation is supplied and otherwise issues a Finish
action. This judgment uses the question and retrieved
evidence, without gold evidence labels at inference.

Constraints are enforced both when constructing the policy
input and before executing its output.
Invalid or immediately repeated outputs trigger
state-dependent recovery, such as reading an unread
candidate or curating inspected candidates.
Raw policy outputs and executed actions are recorded
separately.
The loop returns the selected evidence set upon an accepted
Finish action, exhaustion of the turn limit, or repeated
unusable outputs.
The turn limit counts executed steps, including the initial
retrieval and controller-issued actions.

\subsection{Training Details}
\label{TrainingDetail}
We trained Qwen3.6-27B using 5,000 samples, comprising 1,000 samples drawn randomly from each of five training datasets.
Training consists of 50 updates of teacher-guided behavioral
cloning followed by 250 updates of on-policy self-distillation.
We update only the LoRA parameters, while keeping the backbone
frozen. The teacher shares the same backbone and is evaluated
with the LoRA adapter disabled.
Table~\ref{tab:hyperparameters} summarizes the main hyperparameters.

\begin{table}[h]
    \centering
    \caption{Main training hyperparameters.}
    \label{tab:hyperparameters}
    \begin{tabular}{lc}
        \toprule
        Hyperparameter & Value \\
        \midrule
        Optimizer & AdamW \\
        Learning rate & $1 \times 10^{-5}$ \\
        LoRA rank / $\alpha$ & 16 / 32 \\
        Sampling temperature / top-$p$ & 0.7 / 0.9 \\
        SOD $\epsilon$ / $\delta$ & $10^{-6}$ / 0.5 \\
        Batch size & 10 \\
        \bottomrule
    \end{tabular}
\end{table}

\paragraph{Behavioral-Cloning Warmup.}
We first initialize tool-use behavior within the search
harness through behavioral cloning, following the use of
supervised initialization in prior on-policy distillation
work~\citep{agarwal2024policy}.
The frozen teacher generates actions conditioned on the
current search context $o_t$ and privileged information $z_t$ (i.e., $a_t \sim \pi(\cdot \mid o_t, z_t)$).
The student learns to predict these actions from $o_t$ alone:
\begin{equation}
\mathcal{L}_{\mathrm{BC}}(\theta)
=
-\frac{1}{|\mathcal{I}_{\mathrm{BC}}(\tau)|}
\sum_{t \in \mathcal{I}_{\mathrm{BC}}(\tau)}
\frac{1}{m_t}
\sum_{\ell=1}^{m_t}
\log \pi_\theta
\left(
a_{t,\ell}
\mid o_t, a_{t,<\ell}
\right).
\label{eq:bc_warmup}
\end{equation}

Here, $\tau$ denotes a teacher-generated search trajectory,
$a_t$ is the teacher's action sequence of length $m_t$,
and $\mathcal{I}_{\mathrm{BC}}(\tau)$ contains teacher
decisions executed without harness corrections.
The objective is averaged over the 10 episodes in each update.

\paragraph{Optimization and Inference.}
Training alternates between sampling student trajectories and performing distillation updates. The teacher remains frozen throughout, and the loss is applied only to action tokens generated by the model. At inference time, the teacher and privileged information are removed, while the same state construction and execution mechanisms are retained. At each step, the student makes decisions based on the current context, while the Harness accumulates evidence, tracks search progress, and carries information across steps until a termination condition is met or the search budget is exhausted.

\section{Prompt}

\begin{instructionbox}{Retrieval and Organization}
\noindent
You are a knowledge-base retrieval agent. Find every document
relevant to the query. Think concisely, decide quickly, and
call tools directly.

\par\smallskip
\noindent
\textbf{Tools:} \texttt{\{available\_tools\}}

\par\smallskip
\noindent
\textbf{Rules:}
\begin{enumerate}
    \setlength{\itemsep}{0.3em}
    \setlength{\parsep}{0pt}
    \setlength{\parskip}{0pt}

    \item
    Follow this loop:
    retrieve $\rightarrow$ select $\rightarrow$ judge the
    current evidence. 

    \item
    After every \texttt{\{retrieval\_tools\}} call, select
    before the next retrieval or finish. 

    \item
    Use short queries, search from multiple angles, and
    never repeat the same query.

    \item
    Preserve the complete evidence chain for multi-hop questions.

    \item
    Never finish with an empty selected set while candidates
    exist. Before finish, verify that the evidence chain is
    complete and select all plausibly relevant evidence.

    \item
    Use \texttt{read\_document} for full text and
    \texttt{grep\_corpus} for exact named entities.

    \item
    For comparison questions, run a short targeted search for
    each named entity before finishing. For other multi-hop
    questions, perform at least one follow-up retrieval or
    link exploration after identifying the first hop.
\end{enumerate}

\par\smallskip
\noindent
\texttt{\{working\_memory\}}

\par\smallskip
\noindent
\textbf{Budget:}
\texttt{\{used\_budget\}}/\allowbreak
\texttt{\{total\_budget\}}\allowbreak
\texttt{\{budget\_warning\}}
\end{instructionbox}

\begin{instructionbox}{Evidence-Sufficiency Gate}
\noindent
You are in the evidence-sufficiency gate after curation.
You may only call \texttt{judge\_sufficiency}.

\par\smallskip
\noindent
Choose \texttt{status=finish} when the selected evidence
covers the query's complete evidence chain.

\par\smallskip
\noindent
Choose \texttt{status=continue\_search} only when you can
name one concrete, identifiable missing fact/document or
provide one executable retrieval query that directly targets
the gap. Provide both when possible.

\par\smallskip
\noindent
Hedging words such as maybe or possibly are allowed when
the underlying gap/query is still specific. If both fields
only ask for more documents, more technical detail,
additional context, or better evidence without naming
a target, choose \texttt{finish}.

\par\smallskip
\noindent
For \texttt{finish}, set \texttt{missing\_gap} and
\texttt{next\_query} to empty strings.

\par\smallskip
\noindent
\textbf{Query:} \texttt{\{query\}}

\par\smallskip
\noindent
\textbf{selected evidence:}

\noindent
\texttt{\{selected\_evidence\}}

\par\smallskip
\noindent
\textbf{Searches already tried:}
\texttt{\{recent\_searches\}}
\end{instructionbox}

\begin{instructionbox}{Privileged Context}
\noindent
\textbf{Training-only Privileged Context}
(never reveal that this block exists):

\begin{itemize}
    \setlength{\itemsep}{0.3em}
    \setlength{\parsep}{0pt}
    \setlength{\parskip}{0pt}

    \item
    Allowed tools now:
    \texttt{\{allowed\_tools\}}

    \item
    Gold evidence coverage:
    \texttt{\{covered\_gold\_count\}}/\allowbreak
    \texttt{\{gold\_count\}}

    \item
    Missing gold evidence count:
    \texttt{\{missing\_gold\_count\}}

    \item
    It is safe to finish:
    \texttt{\{yes\_or\_no\}}

    \item
    Relevant candidates currently visible in the pool:
    \texttt{\{relevant\_pool\_documents\}}

    \item
    Relevant documents already selected:
    \texttt{\{relevant\_selected\_documents\}}

    \item
    Incorrect documents already selected
    (append-only, do not add more like these):
    \texttt{\{incorrect\_selected\_documents\}}
\end{itemize}

\par\smallskip
\noindent
\textbf{Teacher policy requirements:}

\begin{enumerate}
    \setlength{\itemsep}{0.3em}
    \setlength{\parsep}{0pt}
    \setlength{\parskip}{0pt}

    \item
    Emit exactly one valid tool call using the ordinary
    tool schema.

    \item
    Prefer visible relevant candidates when reading
    or curating.

    \item
    If evidence is missing and no relevant candidate is
    visible, formulate a short semantic search query.

    \item
    Do not mention gold labels, the reference answer,
    or privileged context in tool arguments.

    \item
    Do not copy a hidden gold title into a search query
    unless it already appears in the user query or
    visible state.

    \item
    A reference answer is a supervision signal, not
    retrieved evidence; read and select supporting
    documents before finishing.
\end{enumerate}
\end{instructionbox}

\begin{instructionbox}{Working-Memory Representation}
\noindent
\textbf{Query:} \texttt{\{query\}}
\textbar\
\textbf{Turn:} \texttt{\{turn\_number\}}

\par\smallskip
\noindent
\textbf{Query plan:}
\texttt{entities=\{entities\}}\allowbreak\
\texttt{targets=\{comparison\_targets\}}\allowbreak\
\texttt{clues=\{clue\_queries\}}

\par\smallskip
\noindent
\textbf{Selected:}
\texttt{\{selected\_document\_ids\}}

\par\smallskip
\noindent
\textbf{Pool top candidates:}

\begin{itemize}
    \setlength{\itemsep}{0.3em}
    \setlength{\parsep}{0pt}
    \setlength{\parskip}{0pt}

    \item
    \texttt{\{document\_id\}}
    \textbar\
    \texttt{\{read\_or\_snippet\}}
    \textbar\
    \texttt{score=\{score\}}
    \textbar\
    \texttt{sources=\{sources\}}

    \par
    \texttt{\{document\_summary\}}
\end{itemize}

\par\smallskip
\noindent
\textbf{Searches tried:}
\texttt{\{recent\_search\_queries\}}

\par\smallskip
\noindent
\textbf{Recent outcomes:}
\texttt{\{recent\_outcomes\}}
\end{instructionbox}


\begin{thebibliography}{39}
\providecommand{\natexlab}[1]{#1}
\providecommand{\url}[1]{\texttt{#1}}
\expandafter\ifx\csname urlstyle\endcsname\relax
  \providecommand{\doi}[1]{doi: #1}\else
  \providecommand{\doi}{doi: \begingroup \urlstyle{rm}\Url}\fi

\bibitem[Agarwal et~al.(2024)Agarwal, Vieillard, Zhou, Stanczyk, Ramos~Garea, Geist, and Bachem]{agarwal2024policy}
Rishabh Agarwal, Nino Vieillard, Yongchao Zhou, Piotr Stanczyk, Sabela Ramos~Garea, Matthieu Geist, and Olivier Bachem.
\newblock On-policy distillation of language models: Learning from self-generated mistakes.
\newblock In \emph{International Conference on Learning Representations}, volume 2024, pp.\  21246--21263, 2024.

\bibitem[Asai et~al.(2024)Asai, Wu, Wang, Sil, and Hajishirzi]{asai2024self}
Akari Asai, Zeqiu Wu, Yizhong Wang, Avi Sil, and Hannaneh Hajishirzi.
\newblock Self-rag: Learning to retrieve, generate, and critique through self-reflection.
\newblock In \emph{International conference on learning representations}, volume 2024, pp.\  9112--9141, 2024.

\bibitem[Chen et~al.(2025)Chen, Li, Sun, Zhou, Zhu, Wang, Pan, Zhang, Chen, Yang, et~al.]{chen2025research}
Mingyang Chen, Tianpeng Li, Haoze Sun, Yijie Zhou, Chenzheng Zhu, Haofen Wang, Jeff~Z Pan, Wen Zhang, Huajun Chen, Fan Yang, et~al.
\newblock Research: Learning to reason with search for llms via reinforcement learning.
\newblock \emph{arXiv preprint arXiv:2503.19470}, 2\penalty0 (3):\penalty0 7, 2025.

\bibitem[Gao et~al.(2025)Gao, Fu, Xie, Xu, He, Mei, Zhu, and Wu]{gao2025beyond}
Jiaxuan Gao, Wei Fu, Minyang Xie, Shusheng Xu, Chuyi He, Zhiyu Mei, Banghua Zhu, and Yi~Wu.
\newblock Beyond ten turns: Unlocking long-horizon agentic search with large-scale asynchronous rl.
\newblock \emph{arXiv preprint arXiv:2508.07976}, 2025.

\bibitem[Guo et~al.(2025)Guo, Yang, Zhang, Song, Wang, Zhu, Xu, Zhang, Ma, Bi, et~al.]{guo2025deepseek}
Daya Guo, Dejian Yang, Haowei Zhang, Junxiao Song, Peiyi Wang, Qihao Zhu, Runxin Xu, Ruoyu Zhang, Shirong Ma, Xiao Bi, et~al.
\newblock Deepseek-r1: Incentivizing reasoning capability in llms via reinforcement learning.
\newblock \emph{arXiv preprint arXiv:2501.12948}, 2025.

\bibitem[He et~al.(2025)He, Huang, Feng, Lin, Zhang, Li, and E]{he-etal-2025-pasa}
Yichen He, Guanhua Huang, Peiyuan Feng, Yuan Lin, Yuchen Zhang, Hang Li, and Weinan E.
\newblock {P}a{S}a: An {LLM} agent for comprehensive academic paper search.
\newblock In \emph{Proceedings of the 63rd Annual Meeting of the Association for Computational Linguistics (Volume 1: Long Papers)}, pp.\  11663--11679, 2025.

\bibitem[Ho et~al.(2020)Ho, Nguyen, Sugawara, and Aizawa]{ho2020constructing}
Xanh Ho, Anh-Khoa~Duong Nguyen, Saku Sugawara, and Akiko Aizawa.
\newblock Constructing a multi-hop qa dataset for comprehensive evaluation of reasoning steps.
\newblock In \emph{Proceedings of the 28th International Conference on Computational Linguistics}, pp.\  6609--6625, 2020.

\bibitem[Izacard \& Grave(2021)Izacard and Grave]{izacard-grave-2021-leveraging}
Gautier Izacard and Edouard Grave.
\newblock Leveraging passage retrieval with generative models for open domain question answering.
\newblock In \emph{Proceedings of the 16th Conference of the European Chapter of the Association for Computational Linguistics: Main Volume}, pp.\  874--880, 2021.

\bibitem[Java et~al.(2026)Java, Koundinyan, Natarajan, and Sharma]{java2026frugalrag}
Abhinav Java, Srivathsan Koundinyan, Nagarajan Natarajan, and Amit Sharma.
\newblock Frugalrag: Less is more in rl finetuning for multi-hop question answering.
\newblock In \emph{International Conference on Learning Representations}, pp.\  146993--147016, 2026.

\bibitem[Jiang et~al.(2023)Jiang, Xu, Gao, Sun, Liu, Dwivedi-Yu, Yang, Callan, and Neubig]{jiang-etal-2023-active}
Zhengbao Jiang, Frank Xu, Luyu Gao, Zhiqing Sun, Qian Liu, Jane Dwivedi-Yu, Yiming Yang, Jamie Callan, and Graham Neubig.
\newblock Active retrieval augmented generation.
\newblock In \emph{Proceedings of the 2023 Conference on Empirical Methods in Natural Language Processing}, pp.\  7969--7992, 2023.

\bibitem[Jin et~al.(2025{\natexlab{a}})Jin, Yoon, Kargupta, Arik, and Han]{jin2025empirical}
Bowen Jin, Jinsung Yoon, Priyanka Kargupta, Sercan~O Arik, and Jiawei Han.
\newblock An empirical study on reinforcement learning for reasoning-search interleaved llm agents.
\newblock \emph{arXiv preprint arXiv:2505.15117}, 2025{\natexlab{a}}.

\bibitem[Jin et~al.(2025{\natexlab{b}})Jin, Zeng, Yue, Yoon, Arik, Wang, Zamani, and Han]{jin2025search}
Bowen Jin, Hansi Zeng, Zhenrui Yue, Jinsung Yoon, Sercan Arik, Dong Wang, Hamed Zamani, and Jiawei Han.
\newblock Search-r1: Training llms to reason and leverage search engines with reinforcement learning.
\newblock \emph{arXiv preprint arXiv:2503.09516}, 2025{\natexlab{b}}.

\bibitem[Joshi et~al.(2017)Joshi, Choi, Weld, and Zettlemoyer]{joshi2017triviaqa}
Mandar Joshi, Eunsol Choi, Daniel~S Weld, and Luke Zettlemoyer.
\newblock Triviaqa: A large scale distantly supervised challenge dataset for reading comprehension.
\newblock In \emph{Proceedings of the 55th Annual Meeting of the Association for Computational Linguistics (Volume 1: Long Papers)}, pp.\  1601--1611, 2017.

\bibitem[Karpukhin et~al.(2020)Karpukhin, Oguz, Min, Lewis, Wu, Edunov, Chen, and Yih]{karpukhin-etal-2020-dense}
Vladimir Karpukhin, Barlas Oguz, Sewon Min, Patrick Lewis, Ledell Wu, Sergey Edunov, Danqi Chen, and Wen-tau Yih.
\newblock Dense passage retrieval for open-domain question answering.
\newblock In \emph{Proceedings of the 2020 Conference on Empirical Methods in Natural Language Processing (EMNLP)}, pp.\  6769--6781, 2020.

\bibitem[Kwiatkowski et~al.(2019)Kwiatkowski, Palomaki, Redfield, Collins, Parikh, Alberti, Epstein, Polosukhin, Devlin, Lee, et~al.]{kwiatkowski2019natural}
Tom Kwiatkowski, Jennimaria Palomaki, Olivia Redfield, Michael Collins, Ankur Parikh, Chris Alberti, Danielle Epstein, Illia Polosukhin, Jacob Devlin, Kenton Lee, et~al.
\newblock Natural questions: a benchmark for question answering research.
\newblock \emph{Transactions of the Association for Computational Linguistics}, 7:\penalty0 453--466, 2019.

\bibitem[Lewis et~al.(2020)Lewis, Perez, Piktus, Petroni, Karpukhin, Goyal, K{\"u}ttler, Lewis, Yih, Rockt{\"a}schel, et~al.]{lewis2020retrieval}
Patrick Lewis, Ethan Perez, Aleksandra Piktus, Fabio Petroni, Vladimir Karpukhin, Naman Goyal, Heinrich K{\"u}ttler, Mike Lewis, Wen-tau Yih, Tim Rockt{\"a}schel, et~al.
\newblock Retrieval-augmented generation for knowledge-intensive nlp tasks.
\newblock \emph{Advances in neural information processing systems}, 33:\penalty0 9459--9474, 2020.

\bibitem[Li et~al.(2026{\natexlab{a}})Li, Jin, Liu, Ding, Wu, Chen, Shen, Qin, Tai, Wang, et~al.]{li2026se}
Jian Li, Yizhang Jin, Dongqi Liu, Hang Ding, Jiafu Wu, Dongsheng Chen, Yunhang Shen, Yulei Qin, Ying Tai, Chengjie Wang, et~al.
\newblock Se-search: Self-evolving search agent via memory and dense reward.
\newblock \emph{arXiv preprint arXiv:2603.03293}, 2026{\natexlab{a}}.

\bibitem[Li et~al.(2025)Li, Dong, Jin, Zhang, Zhou, Zhu, Zhang, and Dou]{li2025search}
Xiaoxi Li, Guanting Dong, Jiajie Jin, Yuyao Zhang, Yujia Zhou, Yutao Zhu, Peitian Zhang, and Zhicheng Dou.
\newblock Search-o1: Agentic search-enhanced large reasoning models.
\newblock In \emph{Proceedings of the 2025 Conference on Empirical Methods in Natural Language Processing}, pp.\  5420--5438, 2025.

\bibitem[Li et~al.(2026{\natexlab{b}})Li, Jin, Dong, Qian, Wu, Wen, Zhu, and Dou]{li2026webthinker}
Xiaoxi Li, Jiajie Jin, Guanting Dong, Hongjin Qian, Yongkang Wu, Ji-Rong Wen, Yutao Zhu, and Zhicheng Dou.
\newblock Webthinker: Empowering large reasoning models with deep research capability.
\newblock \emph{Advances in Neural Information Processing Systems}, 38:\penalty0 120091--120131, 2026{\natexlab{b}}.

\bibitem[Li et~al.(2026{\natexlab{c}})Li, Zhang, Han, Liu, Xie, Zhang, Choi, Zou, and Lu]{li2026flow}
Zhuofeng Li, Haoxiang Zhang, Seungju Han, Sheng Liu, Jianwen Xie, Yu~Zhang, Yejin Choi, James~Y Zou, and Pan Lu.
\newblock In-the-flow agentic system optimization for effective planning and tool use.
\newblock In \emph{International Conference on Learning Representations}, volume 2026, pp.\  50524--50570, 2026{\natexlab{c}}.

\bibitem[Liu et~al.(2024)Liu, Lin, Hewitt, Paranjape, Bevilacqua, Petroni, and Liang]{liu2024lost}
Nelson~F Liu, Kevin Lin, John Hewitt, Ashwin Paranjape, Michele Bevilacqua, Fabio Petroni, and Percy Liang.
\newblock Lost in the middle: How language models use long contexts.
\newblock \emph{Transactions of the association for computational linguistics}, 12:\penalty0 157--173, 2024.

\bibitem[Nakano et~al.(2021)Nakano, Hilton, Balaji, Wu, Ouyang, Kim, Hesse, Jain, Kosaraju, Saunders, et~al.]{nakano2021webgpt}
Reiichiro Nakano, Jacob Hilton, Suchir Balaji, Jeff Wu, Long Ouyang, Christina Kim, Christopher Hesse, Shantanu Jain, Vineet Kosaraju, William Saunders, et~al.
\newblock Webgpt: Browser-assisted question-answering with human feedback.
\newblock \emph{arXiv preprint arXiv:2112.09332}, 2021.

\bibitem[Packer et~al.(2023)Packer, Wooders, Lin, Fang, Patil, Stoica, and Gonzalez]{packer2023memgpt}
Charles Packer, Sarah Wooders, Kevin Lin, Vivian Fang, Shishir~G Patil, Ion Stoica, and Joseph~E Gonzalez.
\newblock Memgpt: Towards llms as operating systems.
\newblock \emph{arXiv preprint arXiv:2310.08560}, 2023.

\bibitem[Press et~al.(2023)Press, Zhang, Min, Schmidt, Smith, and Lewis]{press2023measuring}
Ofir Press, Muru Zhang, Sewon Min, Ludwig Schmidt, Noah~A Smith, and Mike Lewis.
\newblock Measuring and narrowing the compositionality gap in language models.
\newblock In \emph{Findings of the Association for Computational Linguistics: EMNLP 2023}, pp.\  5687--5711, 2023.

\bibitem[{Qwen Team}(2026)]{qwen3.6-27b}
{Qwen Team}.
\newblock {Qwen3.6-27B}: Flagship-level coding in a {27B} dense model, April 2026.
\newblock URL \url{https://qwen.ai/blog?id=qwen3.6-27b}.

\bibitem[Shi et~al.(2026)Shi, Yan, Yin, Verberne, Rijke, and Ren]{shi2026iterative}
Zhengliang Shi, Lingyong Yan, Dawei Yin, Suzan Verberne, Maarten Rijke, and Zhaochun Ren.
\newblock Iterative self-incentivization empowers large language models as agentic searchers.
\newblock \emph{Advances in Neural Information Processing Systems}, 38:\penalty0 135679--135726, 2026.

\bibitem[Song et~al.(2025)Song, Jiang, Min, Chen, Chen, Zhao, Fang, and Wen]{song2025r1}
Huatong Song, Jinhao Jiang, Yingqian Min, Jie Chen, Zhipeng Chen, Wayne~Xin Zhao, Lei Fang, and Ji-Rong Wen.
\newblock R1-searcher: Incentivizing the search capability in llms via reinforcement learning.
\newblock \emph{arXiv preprint arXiv:2503.05592}, 2025.

\bibitem[Trivedi et~al.(2022)Trivedi, Balasubramanian, Khot, and Sabharwal]{trivedi2022musique}
Harsh Trivedi, Niranjan Balasubramanian, Tushar Khot, and Ashish Sabharwal.
\newblock musique: Multihop questions via single-hop question composition.
\newblock \emph{Transactions of the Association for Computational Linguistics}, 10:\penalty0 539--554, 2022.

\bibitem[Trivedi et~al.(2023)Trivedi, Balasubramanian, Khot, and Sabharwal]{trivedi2023interleaving}
Harsh Trivedi, Niranjan Balasubramanian, Tushar Khot, and Ashish Sabharwal.
\newblock Interleaving retrieval with chain-of-thought reasoning for knowledge-intensive multi-step questions.
\newblock In \emph{Proceedings of the 61st annual meeting of the association for computational linguistics (volume 1: long papers)}, pp.\  10014--10037, 2023.

\bibitem[Wang et~al.(2026{\natexlab{a}})Wang, Lian, Zhao, Lin, Chen, and Wang]{wang2026multi}
Jiayuan Wang, Jie Lian, Fu~Zhao, Zhenghong Lin, Hongrong Chen, and Shiping Wang.
\newblock Multi-modal hierarchical retrieval-augmented generation for document question answering.
\newblock In \emph{Proceedings of the 32nd ACM SIGKDD Conference on Knowledge Discovery and Data Mining}, pp.\  12243--12254, 2026{\natexlab{a}}.

\bibitem[Wang et~al.(2026{\natexlab{b}})Wang, Chen, Yang, Huang, Dou, and Wei]{wang2026chain}
Liang Wang, Haonan Chen, Nan Yang, Xiaolong Huang, Zhicheng Dou, and Furu Wei.
\newblock Chain-of-retrieval augmented generation.
\newblock \emph{Advances in Neural Information Processing Systems}, 38:\penalty0 59888--59915, 2026{\natexlab{b}}.

\bibitem[Wu et~al.(2026)Wu, Li, Fang, Yin, Zhang, Wang, Tao, Zhang, Xi, Tang, et~al.]{wu2026webdancer}
Jialong Wu, Baixuan Li, Runnan Fang, Wenbiao Yin, Liwen Zhang, Zhenglin Wang, Zhengwei Tao, Ding-Chu Zhang, Zekun Xi, Robert Tang, et~al.
\newblock Webdancer: Towards autonomous information seeking agency.
\newblock \emph{Advances in Neural Information Processing Systems}, 38:\penalty0 120957--120985, 2026.

\bibitem[Yang et~al.(2018)Yang, Qi, Zhang, Bengio, Cohen, Salakhutdinov, and Manning]{yang2018hotpotqa}
Zhilin Yang, Peng Qi, Saizheng Zhang, Yoshua Bengio, William Cohen, Ruslan Salakhutdinov, and Christopher~D Manning.
\newblock Hotpotqa: A dataset for diverse, explainable multi-hop question answering.
\newblock In \emph{Proceedings of the 2018 conference on empirical methods in natural language processing}, pp.\  2369--2380, 2018.

\bibitem[Yao et~al.(2022)Yao, Zhao, Yu, Du, Shafran, Narasimhan, and Cao]{yao2022react}
Shunyu Yao, Jeffrey Zhao, Dian Yu, Nan Du, Izhak Shafran, Karthik Narasimhan, and Yuan Cao.
\newblock React: Synergizing reasoning and acting in language models.
\newblock \emph{arXiv preprint arXiv:2210.03629}, 2022.

\bibitem[Yuan et~al.(2026)Yuan, Lou, Li, Chen, Lu, Lin, Sun, Zhang, and Han]{yuan-etal-2026-memsearcher}
Qianhao Yuan, Jie Lou, Zichao Li, Jiawei Chen, Yaojie Lu, Hongyu Lin, Le~Sun, Debing Zhang, and Xianpei Han.
\newblock {M}em{S}earcher: Iterative memory integration for search agent via end-to-end reinforcement learning.
\newblock In \emph{Findings of the {A}ssociation for {C}omputational {L}inguistics: {ACL} 2026}, pp.\  14965--14977, 2026.

\bibitem[Zhang et~al.(2025)Zhang, Li, Long, Zhang, Lin, Yang, Xie, Yang, Liu, Lin, Huang, and Zhou]{qwen3embedding}
Yanzhao Zhang, Mingxin Li, Dingkun Long, Xin Zhang, Huan Lin, Baosong Yang, Pengjun Xie, An~Yang, Dayiheng Liu, Junyang Lin, Fei Huang, and Jingren Zhou.
\newblock Qwen3 embedding: Advancing text embedding and reranking through foundation models.
\newblock \emph{arXiv preprint arXiv:2506.05176}, 2025.

\bibitem[Zhao et~al.(2026)Zhao, Xie, Liu, Huang, Pang, Chen, and Grover]{zhao2026self}
Siyan Zhao, Zhihui Xie, Mengchen Liu, Jing Huang, Guan Pang, Feiyu Chen, and Aditya Grover.
\newblock Self-distilled reasoner: On-policy self-distillation for large language models.
\newblock \emph{arXiv preprint arXiv:2601.18734}, 2026.

\bibitem[Zheng et~al.(2025)Zheng, Fu, Hu, Cai, Ye, Lu, and Liu]{zheng-etal-2025-deepresearcher}
Yuxiang Zheng, Dayuan Fu, Xiangkun Hu, Xiaojie Cai, Lyumanshan Ye, Pengrui Lu, and Pengfei Liu.
\newblock {D}eep{R}esearcher: Scaling deep research via reinforcement learning in real-world environments.
\newblock In \emph{Proceedings of the 2025 Conference on Empirical Methods in Natural Language Processing}, pp.\  414--431, 2025.

\bibitem[Zhong et~al.(2026)Zhong, Zheng, Song, Lin, Sun, Jiang, Wang, and Fang]{zhong2026sod}
Qiyong Zhong, Mao Zheng, Mingyang Song, Xin Lin, Jie Sun, Houcheng Jiang, Xiang Wang, and Junfeng Fang.
\newblock Sod: Step-wise on-policy distillation for small language model agents.
\newblock \emph{arXiv preprint arXiv:2605.07725}, 2026.

\end{thebibliography}
\end{document}